\documentclass[10pt]{article} 
\usepackage[preprint]{tmlr}

\usepackage{amsmath,amsfonts,bm}

\def\eqref#1{equation~\ref{#1}}

\def\1{\bm{1}}

\DeclareMathAlphabet{\mathsfit}{\encodingdefault}{\sfdefault}{m}{sl}
\SetMathAlphabet{\mathsfit}{bold}{\encodingdefault}{\sfdefault}{bx}{n}

\usepackage{hyperref}
\usepackage{url}

\usepackage{amssymb}
\usepackage{booktabs}
\usepackage{graphicx}
\usepackage{pifont}
\usepackage{xcolor}

\usepackage[ruled,vlined]{algorithm2e}
\newcommand{\blockcomment}[1]{\bluecommfont{/* #1 */}\;}
\newcommand{\bluecommfont}[1]{\textcolor{blue}{\textit{#1}}}
\SetCommentSty{bluecommfont}
\SetArgSty{textup}      
\DontPrintSemicolon     
\usepackage{colortbl}
\usepackage{array}
\usepackage{makecell}

\definecolor{DarkGreen}{RGB}{0,100,0}
\definecolor{Maroon}{RGB}{128,0,0}

\newcommand{\cmark}{\textcolor{DarkGreen}{\ding{51}}}  
\newcommand{\xmark}{\textcolor{Maroon}{\ding{55}}}      
\newcommand{\pmark}{\textcolor{orange!80!black}{$\sim$}}

\title{FedSSMCoOp: SSM Encoders for light-weight Federated Prompt Learning for Few-shot Classification}

\author{\name Ankita Das \email cs23resch01003@iith.ac.in \\
      \addr Department of Computer Science and Engineering\\
      Indian Institute of Technology Hyderabad
      \AND
      \name Ambarish Parthasarathy \email ai23resch01001@iith.ac.in \\
      \addr Department of Artificial Intelligence\\
      Indian Institute of Technology Hyderabad
      \AND
      \name Sumohana S. Channappayya \email sumohana@ee.iith.ac.in \\
      \addr Department of Electrical Engineering\\
      Indian Institute of Technology Hyderabad
      \AND
      \name C. Krishna Mohan \email ckm@cse.iith.ac.in\\
      \addr Department of Computer Science and Engineering\\
      Indian Institute of Technology Hyderabad}

\begin{document}

\maketitle

\begin{abstract}
Vision-Language Models (VLMs) have shown strong performance across a wide range of downstream vision tasks, thanks to the complementary information contained in the respective domains. Despite the performance gains, most of these approaches rely on aligning these domains using the cosine similarity metric, which fails to capture token-level structure and cross-modal interactions prior to the classification stage. This is especially critical in biomedical applications under federated constraints, where data sharing is restricted, labeled data is scarce at each site, and it differs widely across institutions, leading to substantial statistical heterogeneity. To overcome this issue, we propose FedSSMCoOp, a federated few-shot image classification framework that enables multimodal learning while preserving data privacy. With the help of the SSM-based Vision Mamba and Cross Mamba blocks, and by optimizing only the soft-prompt and communication-prompt updates in the federated setting, the framework prioritizes both computation and performance. Importantly, this eliminates the need to use an external Large Language Model (LLM) for feature alignment. The framework is further trained and evaluated on various biomedical image datasets, and its performance is assessed. The proposed framework delivers stable performance relative to the baselines and is, on average, \textbf{1.96} times lighter. The corresponding script will be made available soon.
\end{abstract}

\section{Introduction}
\label{sec:intro}
The VLMs are a performance-oriented framework where information is processed using one or more complementary modalities. One of the most commonly used modalities is vision and text, where either of them can aid the other.
The features from each of the modalities are either combined in a common space or processed by a dedicated encoder \cite{tan-bansal-2019-lxmert}. Commonly used encoders are attention-based transformers that prioritize performance. The Contrastive Language Image Pretraining (CLIP) is one such VLM that uses this approach in the vision and text domain, respectively \cite{radford2021learning}. The features from both modalities are aligned by means of a contrastive learning approach inspired by InfoNCE \cite{oord2018representation}. This is the foundational template for most of the present-day VLMs. On the other hand, the Bootstrapped Language Image Pretraining (BLIP), the second variant of the prior CLIP model, is designed for both retrieval and generative applications \cite{li2022blip}. Further, the BLIP introduces a filtering mechanism called \textit{CaptFilt} that filters out noisy captions, thereby making it more robust to perturbations. The SigLIP \cite{zhai2023sigmoid} is another variant of the CLIP model in which the contrastive approach is replaced with a sigmoid objective, which is simple but has proven quite effective in terms of performance and stability. Biomedical imaging is an important application domain for VLMs because the data is typically distributed across multiple healthcare institutions and cannot be directly shared due to privacy constraints. In addition, biomedical images can differ substantially across institutions due to variations in imaging devices, acquisition protocols, and patient populations. These characteristics make biomedical imaging a challenging and highly heterogeneous setting for VLM adaptation, particularly when only a small number of labeled samples are available. \\
Despite the advancements, the transformer-based VLMs are computationally inefficient due to the presence of an attention mechanism. The State Space Modeling (SSMs) of Mamba \cite{gu2023mamba} was effective in reducing the computation complexity from quadratic in attention to almost linear. This helped to curb the computational complexity while simultaneously maintaining performance, and is widely used in most of the applications in Biomedical and Satellite Imaging, respectively. The Vision Mamba \cite{zhu2024vision} is an extension of this approach in the vision domain, using a bidirectional state-space approach for an effective scanning mechanism. The VMamba \cite{liu2024vmamba}, on the other hand, used an unidirectional stage-wise approach, mimicking an SSM-based mixer model.\\
However, every advancement comes at a cost. The Mamba variants, despite being lightweight, weren't as effective as attention-based models. This is because the attention mechanism was effective at extracting fine-grained features, whereas Mamba focuses more on the global perspective of features \cite{parthasarathy2026vimclip}. These factors are also crucial in the federated environments, where the performance and the latency between the client and the server must be jointly optimized. Given that the data across participating clients are typically non-independent and identically distributed (non-IID), each client's data is drawn from a different underlying distribution, which further increases the difficulty of achieving stable convergence. These characteristics create a unique trade-off in federated biomedical VLMs: the model should be computationally efficient enough to communicate across clients, while still being capable of capturing both global and fine-grained features from heterogeneous medical images. Therefore, the choice of the visual encoder and the strategy for adapting it are particularly important in this setting. The computational efficiency of SSM-based models is particularly attractive here, where both local computation and communication between clients and the server are important. Therefore, we propose \textbf{FedSSMCoOp}, a federated framework that integrates Vision Mamba with the Cross-Mamba Interaction block \cite{liu2026cross, zhu2024vision}, balancing performance and computational complexity. The integrated design is further stitched into a Federated Learning (FL) setup that communicates only with the soft prompt deltas. This reduces the per-round communication cost between the client and the server. We demonstrate gains and on-par performance across several datasets while significantly reducing computational cost. Further, this approach is trained and validated across biomedical imaging datasets and compared over the baselines under the non-IID data distributions. 

\section{Related Works} \label{sec:relworks}
\begin{table}[h!]
\centering
\caption{Comparison of related methods with FedSSMCoOp.
\textbf{FL}: Federated Learning, \textbf{PL}: Prompt Learning,
\textbf{Vim}: Vision Mamba encoder, \textbf{MIB}: Mamba Interaction Block.
\cmark: supported, \xmark: not supported, \pmark: partially supported.}
\label{tab:related_works}
\begin{tabular}{lcccccc}
\toprule
\hline
\textbf{Method} & \textbf{FL} & \textbf{PL} & \textbf{Vim} & \textbf{MIB} & \textbf{Medical} & \textbf{Few-shot} \\
\hline
\multicolumn{7}{l}{\textit{Federated Learning}} \\ \hline
FedAvg\cite{mcmahan2017communication}
& \cmark & \xmark & \xmark & \xmark & \pmark & \xmark \\
FedProx\cite{li2020federated}
& \cmark & \xmark & \xmark & \xmark & \pmark & \xmark \\
SCAFFOLD\cite{karimireddy2020scaffold}
& \cmark & \xmark & \xmark & \xmark & \xmark & \xmark \\
\hline
\multicolumn{7}{l}{\textit{Prompt Learning}} \\ \hline
CoOp\cite{zhou2022learning}
& \xmark & \cmark & \xmark & \xmark & \xmark & \cmark \\
CoCoOp\cite{zhou2022conditional}
& \xmark & \cmark & \xmark & \xmark & \xmark & \cmark \\
KgCoOp\cite{yao2023visual}
& \xmark & \cmark & \xmark & \xmark & \xmark & \cmark \\
ProDA\cite{lu2022prompt}
& \xmark & \cmark & \xmark & \xmark & \xmark & \cmark \\
PromptFL\cite{guo2023promptfl}
& \cmark & \cmark & \xmark & \xmark & \xmark & \cmark \\
BiomedCoOp\cite{koleilat2025biomedcoop}
& \xmark & \cmark & \xmark & \xmark & \cmark & \cmark \\
FedTPG\cite{qiu2024federated}
& \cmark & \cmark & \xmark & \xmark & \xmark & \cmark \\
FedMVP\cite{singha2025fedmvp}
& \cmark & \cmark & \xmark & \xmark & \xmark & \cmark \\
FedMPT\cite{wang2026fedmpt}
& \cmark & \cmark & \xmark & \xmark & \xmark & \cmark \\
BioVLM\cite{singha2026biovlm}
& \xmark & \cmark & \xmark & \xmark & \cmark & \cmark \\
\hline
\multicolumn{7}{l}{\textit{State Space Models}} \\ \hline
Mamba\cite{gu2023mamba}
& \xmark & \xmark & \xmark & \xmark & \xmark & \xmark \\
Vision Mamba\cite{zhu2024vision}
& \xmark & \xmark & \cmark & \xmark & \xmark & \xmark \\
VMamba\cite{liu2024vmamba}
& \xmark & \xmark & \cmark & \xmark & \xmark & \xmark \\
MedMamba\cite{yue2024medmamba}
& \xmark & \xmark & \cmark & \xmark & \cmark & \xmark \\
MambaDFuse\cite{li2024mambadfuse}
& \xmark & \xmark & \cmark & \cmark & \xmark & \xmark \\
COMO\cite{liu2026cross}
& \xmark & \xmark & \cmark & \cmark & \xmark & \xmark \\
\hline
\textbf{FedSSMCoOp (ours)}
& \cmark & \cmark & \cmark & \cmark & \cmark & \cmark \\
\hline
\bottomrule
\end{tabular}
\end{table}
Most previous work focused on either the \textit{Federated Learning} or the \textit{Prompt Learning}. Recent approaches use LLMs to leverage their performance and report few-shot results for better generalization. A similar trend was also observed for the SSMs, where the results were reported mainly for unimodal, application-oriented approaches. 
Federated learning is quite effective for most tasks, enabling collaborative training without explicitly sharing information. Keeping the regulations like HIPAA and GDPR into consideration, the FedAvg \cite{mcmahan2017communication} focuses on a decentralized mechanism across sequential models, particularly on mobile devices. On the other hand, FedProx \cite{li2020federated} and SCAFFOLD \cite{karimireddy2020scaffold} address non-IID heterogeneity and share full model updates within the server. But this approach increases per-round communication costs and leads to unstable gradient inversion. \\ Further, in the domain of \textit{Prompt Learning}, CoOp \cite{zhou2022learning} helps to address the issue of communication cost by optimizing the learnable context tokens, which can be adapted from a frozen CLIP encoder. Works like CoCoOp \cite{zhou2022conditional} and KgCoOp \cite{yao2023visual} use probabilistic conditioning and further compute the cosine similarity between the soft and hard prompts through an external knowledge mechanism. Further, PromptFL \cite{guo2023pfedprompt}, aggregates only the soft prompts, and the BiomedCoOp \cite{koleilat2025biomedcoop} builds on top of this approach through an LLM-generated class descriptions and semantic prototypes. FedTPG \cite{qiu2024federated} implemented a shared text-conditioned prompt generator in this framework. To better understand and learn the multimodal prompts, FedMVP \cite{singha2025fedmvp} combines the vision features with LLM-derived class attributes. As an extension of this the FedMPT \cite{wang2026fedmpt} uses this LLM-based class attributes to improve federated multi-label recognition. In biomedical domain BioVLM \cite{singha2026biovlm} learns a dynamic prompt bank by distilling LLM-derived attributes for few-shot biomedical classification.\\
The introduction of SSMs such as Mamba \cite{gu2023mamba} provides a lightweight alternative to the bulky transformer-based models. The Vision Mamba \cite{zhu2024vision} adapts a similar mechanism to the vision domain by serializing patches into a bidirectional SSMs approach, one for the forward pass and the other for the backward pass. The VMamba\cite{liu2024vmamba} incorporates a stage-wise scanning approach similar to a Mixer model. The MedMamba \cite{yue2024medmamba} uses the strengths both the Vision Mamba and convolution blocks for medical image classification. On the other hand, MambaDFuse \cite{li2024mambadfuse} and COMO \cite{liu2026cross} introduce dual-stream Mamba architectures that fuse two feature streams and serve as an alternative to the bulky cross attention mechanism. \\
However, these existing prompt learning methods mainly operate in centralized setting or on the cosine similarity between the two modalities don't necessarily capture the cross-modal interaction. Further, most of the Mamba-based methods mainly rely on unimodal setups, and fewer works contribute to multimodal and federated setups. To the best of our knowledge, the proposed FedSSMCoOp is a work that addresses these issues, integrating the federated setup with SSM approaches of Vision Mamba and Cross Mamba block (COMO), capturing cross-modal cues for effective alignment. This eliminates the use of bulky LLMs, as the COMO already solves this. The comparison of the different approaches with reference to the proposed framework is shown in Table \ref{tab:related_works}. 
\section{Preliminaries}
\subsection{State Space Models (SSMs)} \label{ssm}
Given an input image $x \in \mathbb{R}^{H \times W \times 3}$, it is first divided into \textit{N} non-overlapping patches, similar to that of the transformer model. Now, each patch is linearly projected to a $d_v$-dimensional embedding, along with  learnable positional embeddings in order to preserve spatial ordering. Vision Mamba models the patch sequence as a selective state space model (SSM) with the following recurrent formulation:
\begin{align} \label{equation4}
    h_t &= \mathbf{A}_t h_{t-1} + \mathbf{B}_t e_t \\
    o_t &= \mathbf{C}_t h_t
\end{align}
where $e_t \in \mathbb{R}^{d_v}$ is the embedding of the $t$-th patch, $h_t \in \mathbb{R}^{d_s}$ is hidden state capturing context from all previously seen patches, and 
$\mathbf{A}_t, \mathbf{B}_t, \mathbf{C}_t$ are input-dependent projection matrices that implement selective gating. Importantly, the matrices $\mathbf{B}_t$ and $\mathbf{C}_t$ are functions of input $e_t$, allowing the model to selectively absorb informative patches into memory while suppressing background patches. Importantly, with reference to Equation \ref{equation4}, it can be observed that it clearly operates in the continuous domain. Since the Mamba operates in the discrete domain, an intermediate parameter $\Delta$ is used to convert the above variables in the discrete space. 
\begin{equation}
\begin{aligned}
\bar{A} &= \exp(\Delta A), \\
\bar{B} &= (\Delta A)^{-1}
          \left(\exp(\Delta A)-I\right)\Delta B, \\
h_t &= \bar{A}h_{t-1} + \bar{B}e_t, \\
y_t &= Ch_t.
\end{aligned}
\label{eq:mamba_discrete}
\end{equation}

\begin{equation}
\begin{aligned}
y &= \left(
\overline{CB},
\overline{CAB},
\ldots,
\overline{C}\,\overline{A}^{M-1}\overline{B}
\right)e_t, \\
y &= K * e.
\end{aligned}
\label{eq:mamba_convolution}
\end{equation}
The output is then computed using the convolution and the sequential mode, helping in both optimized training and inference. Here, $M$ is the sequence length of the input, and $K$ is the corresponding kernel having a dimension of $\overline{K} \in \mathbb{R}^{M}$.

Further, in the case of the cross-modal mechanism, the COMO \cite{liu2026cross} is a much lighter alternative to the attention-based mechanism. It gets its inspiration from the dual-streamed architecture and combines the spatial features from the two modalities by means of a Cross mamba Block \cite{liu2026cross}. Firstly, the high-level features, a combination of both average and max pooling, are applied along with deep feature mapping in order to make the model more adaptive. These processed features $S_{in} \in \mathbb{R}^{HW X C}$ are further flattened, and are incorporated with a learning position embedding $P \in \mathbb{R}^{HW X C}$, providing the necessary positional encodings. Now, the $S_{in}$ is scanned across the four directions ($CRScan_{i}$), helping in the expansion of the serialization approach. The results of this, goes across the SSM structure (\textbf{S6}), further resulting in four separate outputs.
\begin{equation} \label{equation-smamba}
\left\{\begin{array}{rl}
x_i & =\operatorname{CRScan_i}\left(S_{i n}\right), \\
y_i & =\mathbf{S 6}_i\left(x_i\right), \\
S_{\text {out }} & =\sum_{i=1}^4 \operatorname{reverse scan}_i\left(y_i\right) .
\end{array} i=1,2,3,4 .\right.
\end{equation}
\\
Similarly, for the Cross-mamba block, the inputs from both modalities are processed. The governing expression is very similar to that of the Equation \ref{equation-smamba}. 
\begin{equation}
\left\{
\begin{array}{rl}
x_i^1, x_i^2 &= \operatorname{CRScan}_i\left(S_{\mathrm{in}}^1,S_{\mathrm{in}}^2\right), \\
y_i &= \mathbf{CS6}_i\left(x_i^1,x_i^2\right), \\
S_{\mathrm{out}} &= \sum_{i=1}^{4} \operatorname{reverse\_scan}_i\left(y_i\right).
\end{array}
\right.
\qquad i=1,2,3,4.
\label{eq:cross_scan}
\end{equation} \\
where, $S_{i n}^1,S_{i n}^2$ are the multimodal inputs for the Cross mamba block, with the \textbf{CS6} being the main structure of the cross mamba, and $x_i^1, x_i^2$ are the serialized inputs from the two modalities.

\subsection{Federated Learning}
Federated Learning \cite{mcmahan2017communication} is a distributed machine learning paradigm in which multiple clients collaboratively train a shared model without exchanging their private data. Considering $K$ clients, where each client $k \in \{1, \ldots, K\}$ has a local dataset $\mathcal{D}_k = \{(x_i, y_i)\}_{i=1}^{n_k}$ which remains private. The objective is to learn a global model with parameter $\theta$ that minimizes the local objectives:
\begin{equation}
    \min_{\theta} \mathcal{L}(\theta) = \sum_{k=1}^K \frac{n_k}{n} \mathcal{L}_k(\theta), \quad n = \sum_{k=1}^K n_k
\end{equation}
where $\mathcal{L}_k(\theta) = \mathbb{E}_{(x,y) \sim \mathcal{D}_k}[\ell(\theta; x, y)]$ is the local loss on client $k$. The FedAvg \cite{mcmahan2017communication} algorithm proceeds in synchronous rounds. At each round the server broadcasts the current global parameters $\theta^{(t)}$ to all the clients and each client $k$ performs $E$ local epochs of stochastic gradient descent on its own private data and obtains updated parameters $\theta_k^{(t+1)}$. Then the server aggregates client updates using weighted averaging:
\begin{equation}
    \theta^{(t+1)} = \sum_{k=1}^K \frac{n_k}{n} \theta_k^{(t+1)}
\end{equation}
Because of this, the raw data never leaves the clients, preserving privacy. Another challenge in FL is data heterogeneity, where datasets in clients follow different data distributions ($\mathcal{D}_k \not\sim \mathcal{D}_j$ for $k \neq j$ a setting known as non-IID). To simulate this in medical imaging, class labels are partitioned across clients using Dirichlet distribution, $\text{Dir}(\alpha)$, where a smaller $\alpha$ induces greater heterogeneity. Communication cost with the server is another challenge, as full model weights for vision encoders can be heavy, making each communication round expensive, especially on bandwidth-constrained hospital networks. Prompt-based federated learning (FL) \cite{guo2023promptfl} addresses this by sharing only a small set of learnable parameters, reducing communication cost while keeping all encoder weights frozen and local.

\section{Proposed Approach}
\begin{figure*}[!ht]
    \caption{The architecture of the proposed FedSSMCoOp framework. On each client, a frozen Vison Mamba encoder produces image tokens $V$ while the text encoder produces text tokens $T$. The COMO block fuses both the streams bidirectionally into text-aware and image-aware features, which are mean-pooled to form $\mathrm{Pr}_{student}$ which is optimized using the cross-entropy loss $\mathcal{L}_{CE}$. Only the prompt parameters are sent to the server for aggregation.}
    \centering
    \includegraphics[width=0.9\textwidth]{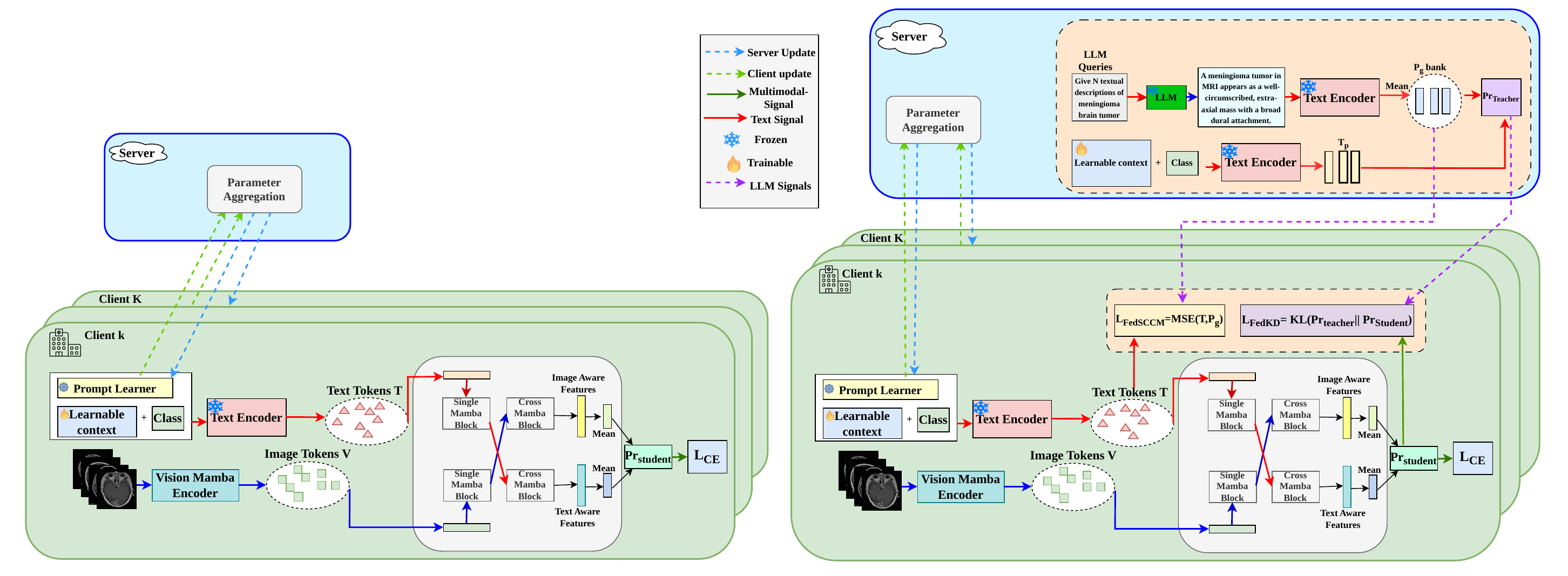}
    \label{fig:FedSSMCoOp_framework}
\end{figure*}

In this work, we introduce the \textbf{FedSSMCoOp} framework, a federated approach to zero-shot and few-shot image classification that exploits a decentralized \textit{Prompt Learning} objective to optimize communication cost between clients and the server. Further, to optimize the computational complexity of the attention mechanism, we first bring in the Vision Mamba \cite{zhu2024vision} at the vision encoder end. Further, to capture the inter-modality cues, we incorporate the Cross Mamba interaction block \cite{liu2026cross}, importantly, to the vision and the text modality, respectively. Since this interaction block already captures and aligns the corresponding modal features, we believe that this can substitute the role of the LLM. The corresponding architecture and the algorithm of the proposed framework is shown in Figure \ref{fig:FedSSMCoOp_framework} and Algorithm \ref{alg:FedSSMCoOp}

\begin{algorithm}[!ht]
\footnotesize
\caption{\textbf{FedSSMCoOp Federated Training}}
\label{alg:FedSSMCoOp}

\KwData{Number of clients $K$, local epochs $E$, global rounds $T$,
learning rate $\eta$, and client datasets $\{\mathcal{D}_k\}_{k=1}^{K}$.}
\KwData{Frozen Vision Mamba encoder $\mathcal{E}_v$ and
BiomedCLIP text encoder $\mathcal{E}_t$.}
\KwResult{Trained global prompt $\bar{P}^{(T)}$ and per-client MIB parameters
$\{\Theta_{\mathrm{MIB}}^k\}_{k=1}^{K}$.}

\blockcomment{Server initialization}
Initialize global prompt $\bar{P}^{(0)} \in \mathbb{R}^{M \times d}$
from ``a photo of a''\;
\For{each client $k \in \{1,\ldots,K\}$}{
    Initialize MIB parameters $\Theta_{\mathrm{MIB}}^k$ and
    text projection $W_{\mathrm{txt}}^k$ locally\;
}
\For{round $t = 0,1,\ldots,T-1$}{
    \blockcomment{Server broadcasts the global prompt}
    Broadcast global prompt $\bar{P}^{(t)}$ to all clients\;
    \For{each client $k \in \{1,\ldots,K\}$ \textbf{in parallel}}{
        $X_p^k \leftarrow \bar{P}^{(t)}$\;
        \For{local epoch $e = 1,\ldots,E$}{
            \For{mini-batch $(x,y) \in \mathcal{D}_k$}{
                $V \leftarrow \mathcal{E}_v(x)$
                \tcc*[r]{frozen image encoder}
                $T \leftarrow \mathcal{E}_t([X_p^k;\texttt{cls}_c]),
                \quad \forall c \in \{1,\ldots,C\}$\;
                $T_{\mathrm{proj}} \leftarrow W_{\mathrm{txt}}^k T$
                \tcc*[r]{local projection}
                $V',T'_{\mathrm{proj}} \leftarrow
                \mathrm{MIB}(V,T_{\mathrm{proj}};\Theta_{\mathrm{MIB}}^k)$
                \tcc*[r]{local MIB}
                $\mathrm{logits} \leftarrow \tau \cdot \mathrm{mean}(V')
                \cdot {T'_{\mathrm{proj}}}^{\top}$\;
                $\Pr_{\mathrm{student}} \leftarrow \mathrm{softmax}(\mathrm{logits})$\;
                $\mathcal{L}_{\mathrm{CE}} \leftarrow
                -\sum_c y_c \log \Pr_{\mathrm{student},c}$\;
                Update $X_p^k$, $W_{\mathrm{txt}}^k$, and
                $\Theta_{\mathrm{MIB}}^k$ via SGD on $\mathcal{L}_{\mathrm{CE}}$\;
            }
        }
        \blockcomment{Compute local prompt update}
        $\Delta X_p^k \leftarrow X_p^k-\bar{P}^{(t)}$\;
        Upload $\Delta X_p^k$ and $n_k$
        \tcc*[r]{only prompt delta is uploaded}
    }
    \blockcomment{Server-side FedAvg aggregation}
    $\bar{P}^{(t+1)} \leftarrow \bar{P}^{(t)} +
    \displaystyle\sum_{k=1}^{K} \frac{n_k}{n}\Delta X_p^k$\;
}
\Return{$\bar{P}^{(T)}$ and $\{\Theta_{\mathrm{MIB}}^k\}_{k=1}^{K}$}
\end{algorithm}

We consider a federated framework with $K$ clients, where each client $k \in \{1, \ldots, K\}$ holds a private dataset $\mathcal{D}_k = \{(x_i, y_i)\}_{i=1}^{n_k}$ of medical images $x_i \in \mathbb{R}^{H \times W \times 3}$ and class labels $y_i \in \{1, \ldots, C\}$. The data remains local and is never shared across the clients or with the central server at any point during training. In the federated few-shot setting, each client trains on at most $K$-shot samples per class, giving a total of $N_{\text{total}} = K \times C$ training images distributed across all clients. The data is partitioned using Dirichlet distribution $\text{Dir}(\alpha)$ to simulate realistic non-IID clinical distributions, where smaller $\alpha$ induces greater heterogeneity across clients. The goal is to learn a classifier that generalizes to all $C$ classes, including classes unseen at individual clients, through $T$ global communication rounds of federated training. Each client maintains a set of learnable soft prompt context tokens $X_p^k \in \mathbb{R}^{M \times d}$, where the subscript $p$ denotes the \emph{prompt} parameters, the superscript $k$ indexes the client, $M$ is the number of context tokens and $d$ is the embedding dimension. Importantly, the server maintains only the global prompt  $\bar{P} \in \mathbb{R}^{M \times d}$  from which each client initializes its local $X_p^k$ at the start of every round. Each round, every client uploads only the prompt delta $\Delta X_p^k = X_p^{k,\text{after}} - \bar{P}$, ensuring that communication cost scales with the prompt size.

Each client computes its predictions using the Vim \cite{zhu2024vision} encoder for image branch and text encoder with learnable soft prompt $X_p^k$ for text branch. The input image $x$ is processed through Vim \cite{zhu2024vision} to obtain the patch tokens $V \in \mathbb{R}^{N \times d_v}$, while the text encoder produces class tokens $T \in \mathbb{R}^{C \times 512}$ from the concatenation $[X_p^k;\, \texttt{cls}_c]$ for each class $c$. the text tokens are projected via a learnable linear layer, giving $T_{\text{proj}} \in \mathbb{R}^{C \times d_v}$. The Mamba Interaction Block (MIB) \cite{liu2026cross} as mentioned in Section \ref{ssm}, fuses the image and text tokens bidirectionally, producing text-aware image patches $V'$ and image-aware class tokens $T'_{\text{proj}}$. The class logits are computed by mean-pooling $V'$ and taking the inner product with $T'_{\text{proj}}$. During the \textit{local training} each client minimizes the standard cross-entropy loss. Only the soft prompt $X_p^k$, the text projection matrix  $W_{\text{txt}}$, and the MIB parameters receive the gradient updates. The Vision Mamba and BiomedCLIP text encoders remain frozen throughout the training process, as shown in Algorithm \ref{alg:FedSSMCoOp}.

At each round $t$, the server broadcasts the current global prompt $\bar{P}^{(t)}$ to all $K$ clients. Each client $k$ initializes $X_p^k \leftarrow \bar{P}^{(t)}$, performs $E$ local epochs using SGD on $\mathcal{L}_{\text{CE}}$, and uploads only the prompt delta $\Delta X_p^k = X_p^{k,\text{after}} - \bar{P}^{(t)}$ along with its local sample count $n_k$. The server aggregates client updates  via weighted averaging \cite{mcmahan2017communication}:
\begin{equation}
    \bar{P}^{(t+1)} = \bar{P}^{(t)} + \sum_{k=1}^K \frac{n_k}{n} \Delta X_p^k
\end{equation} 
This process follows the prompt-based federated paradigm introduced by PromptFL \cite{guo2023pfedprompt}, in which only the lightweight prompt parameters are communicated between clients and the server. Since $\Delta X_p^k$ is soft weight the per-round communication cost is reduced compared to full Vision Mamba weight sharing. As the raw data never leave clients and only prompt deltas are transmitted to the server, preserving privacy. It is important to note that only the soft prompts $\bar{p}$ are aggregated at the server. The text projection matrix $W_{\text{txt}}^k$ and the MIB parameters $\Theta_{\text{MIB}}^k$ are optimized locally and remain client-specific throughout both training and inference; they are never uploaded or aggregated. The shared prompt $\bar{P}$ captures knowledge across clients, while the client-specific $W_{\text{txt}}^k$ and $\Theta_{\text{MIB}}^k$ adapt the model to each client's non-IID data distribution. During inference time client $k$ performs prediction using the aggregated global prompt $\bar{P}^{(T)}$ together with its locally learned $W_{\text{txt}}^k$ and $\Theta_{\text{MIB}}^k$, and we report the mean performance across clients. By aggregating only the soft prompt, FedSSMCoOp substantially reduces communication overhead as reported in Section \ref{tab:efficiency-all}.
\section{Experiments}
In the case of the experimental setup, the models such as BiomedCoOp \cite{koleilat2025biomedcoop}, CoCoOP \cite{zhou2022conditional}, KgCoOp \cite{yao2023visual} models respectively, were adopted for the Prompt Learning objective. In this work, we extend these models to the federated setting and keep this as the baseline, and compare our proposed FedSSMCoOp framework. 
In order to evaluate the performance of our framework, we evaluate on 11 diverse biomedical datasets, including Magnetic Resonance Imaging (MRI),  Ultrasound, histopathology, X-ray Imaging, Optical Coherence Tomography (OCT), gastrointestinal endoscopy, dermatoscopy, fundus imaging, and computed tomography (CT). The descriptions of the dataset are shown in Table \ref{tab:dataset_overview}. These datasets are particularly chosen, as they exhibit properties that make the federated prompt-learning setup particularly challenging.
\begin{table*}[h!]
\centering
\caption{Overview of the eleven biomedical imaging datasets used in our evaluation.}
\setlength{\tabcolsep}{4pt}
\renewcommand{\arraystretch}{1.1}
\scriptsize
\resizebox{\textwidth}{!}{%
\begin{tabular}{@{}l | l l c p{6.2cm} c@{}}
\toprule
\textbf{Dataset} & \textbf{Modality} & \textbf{Organ} & \textbf{\#Cls} & \textbf{Classes} & \textbf{Train / Test / Val} \\
\midrule
BTMRI\textsuperscript{1}                              & Brain MRI      & Brain         & 4 & Glioma, Meningioma, Normal, Pituitary & 2854 / 1717 / 1141 \\
\hline
BUSI~\cite{al2020dataset}                             & Breast US      & Breast        & 3 & Benign, Malignant, Normal & 389 / 236 / 155 \\
\hline
CHMNIST~\cite{kather2016multi}                        & Histopathology & Colon         & 8 & Adipose, Complex Stroma, Debris, Empty, Immune, Normal Mucosa, Simple Stroma, Tumor Epithelium & 2496 / 1504 / 1000 \\
\hline
COVID~\cite{tahir2021covid}                           & Chest X-ray    & Lungs         & 4 & COVID-19, Lung Opacity, Normal, Viral Pneumonia & 10582 / 6351 / 4232 \\
\hline
LC25000~\cite{borkowski1912lung}                      & Histopathology & Lung \& Colon & 5 & Colon Adeno., Colon Benign, Lung Adeno., Lung Benign, Lung SCC & 12500 / 7500 / 5000 \\
\hline
OCTMNIST~\cite{kermany2018identifying}                & Retinal OCT    & Retina        & 4 & CNV, Drusen, DME, Normal & 97477 / 1000 / 10832 \\
\hline
Kvasir~\cite{pogorelov2017kvasir}                     & GI Endoscopy   & GI Tract      & 8 & Dyed Lifted Polyps, Cecum, Esophagitis, Dyed Resection, Pylorus, Z-Line, Polyps, Ulcerative Colitis & 2000 / 1200 / 800 \\
DermaMNIST~\cite{codella2019skin,tschandl2018ham10000} & Dermatoscopy  & Skin          & 7 & Actinic Keratosis, BCC, Benign Keratosis, Dermatofibroma, Nevus, Melanoma, Vascular Lesion & 7007 / 2005 / 1003 \\
\hline
RetinaMNIST~\cite{kohler2013automatic}                & Fundus         & Retina        & 4 & Cataract, Diabetic Retinopathy, Glaucoma, Normal & 2108 / 1268 / 841 \\
\hline
CTKidney~\cite{islam2022vision}                       & CT Scan        & Kidney        & 4 & Cyst, Stone, Tumor, Normal & 6221 / 3738 / 2487 \\
\hline
KneeXray~\cite{chen2018knee}                          & X-ray          & Knee          & 5 & None, Doubtful, Minimal, Moderate, Severe (OA grade) & 5778 / 1656 / 826 \\
\bottomrule
\end{tabular}%
}
\label{tab:dataset_overview}
\end{table*}

\begin{table*}[h!]
\centering
\caption{The corresponding modality-wise parameters of the models. Note that the PromptFL framework makes use of a ResNet-based model in the vision modality.}
\label{tab:parameters-comparison}
\resizebox{\textwidth}{!}{%
\begin{tabular}{l||c|c|c|c|c|c|c||c}
\hline
\textbf{Model/Modality} & \textbf{BioMedCLIP} & \textbf{CoCoOp} & \textbf{KgCoOp} & \textbf{PromptFL} & \textbf{PromptFL+FedProx} & \textbf{FedOTP} & \textbf{BioMedCoOp} & \textbf{FedSSMCoOp} \\ \hline
Vision Modality & 86M & 86M & 86M & 38M & 38M & 38M & 86M & 26M \\ \hline
Text Modality & 110M & 38M & 38M & 38M & 38M & 38M & 38M & 38M \\ \hline
Others & - & Prompt (35k) & Prompt (659k) & Prompt (2k) & Prompt (2k) & Prompt (4k) & Prompt (86M) & Cross-Mamba (5M) \\ \hline
Total Parameters & 196M & 124M & 125M & 77M & 77M & 77M & 210M & 69M \\ \hline
\end{tabular}%
}
\end{table*}
In this framework, we simulate the federated learning setup with clients $\{4, 8, 16\}$ under non-IID data distributions, and with a Dirichlet partitioning strategy, $\alpha \in \{0.1, 0.5, 1.0\}$, where smaller $\alpha$ induces higher heterogeneity. For all experiments, training was performed for 100 communication rounds, with 5 local epochs per communication round. In this paper, we report results with $8$ clients and $\alpha=0.5$, representing a moderate heterogeneity setting. The few-shot classification performance is reported here, while the results of the zero-shot classification are provided in the Appendix. The results are reported on the test set and averaged over three random seed values. Importantly, at the time of training, only the lightweight soft prompts are updates and communicated between clients and the server. The server aggregates the updates from the clients using the weighted average technique, while prototype updates are combined using the trimmed mean for robustness.

\begin{table*}[h!]
\centering
\caption{The quantitative few-shot classification accuracy ($k=1$) results on the respective baselines and the proposed framework on different datasets. For each dataset, \textbf{best-performing} and \textcolor{blue!70!black}{\underline{second-best performing}} models are reported.}
\label{tab:main_results}

\resizebox{\textwidth}{!}{%
\begin{tabular}{lcccccccccc||c}
\toprule
\textbf{Dataset}
& \textbf{FedBiomedCLIP}
& \textbf{FedCoCoOp}
& \textbf{FedKgCoOp}
& \textbf{PromptFL}
& \textbf{PromptFL+FedProx}
& \textbf{FedOTP}
& \textbf{FedTPG}
& \textbf{FedMVP}
& \textbf{FedMPT}
& \textbf{FedBiomedCoOp}
& \textbf{FedSSMCoOp (Ours)}\\
\midrule

BTMRI
& \textbf{57.08{\scalebox{0.55}{$\pm$0.00}}}
& 48.86{\scalebox{0.55}{$\pm$7.92}}
& 51.19{\scalebox{0.55}{$\pm$4.62}}
& 41.60{\scalebox{0.55}{$\pm$3.29}}
& 36.93{\scalebox{0.55}{$\pm$5.44}}
& 31.43{\scalebox{0.55}{$\pm$3.93}}
& 47.13{\scalebox{0.55}{$\pm$10.10}}
& 44.321
& 23.12
& 50.67{\scalebox{0.55}{$\pm$4.74}}
& \textcolor{blue!70!black}{\underline{53.64{\scalebox{0.55}{$\pm$1.78}}}} \\

BUSI
& \textcolor{blue!70!black}{\underline{51.27{\scalebox{0.55}{$\pm$0.00}}}}
& 43.22{\scalebox{0.55}{$\pm$6.27}}
& 39.69{\scalebox{0.55}{$\pm$4.73}}
& 43.30{\scalebox{0.55}{$\pm$12.72}}
& 37.43{\scalebox{0.55}{$\pm$12.65}}
& 33.33{\scalebox{0.55}{$\pm$6.91}}
& 37.57{\scalebox{0.55}{$\pm$7.95}}
& 40.678
& \textbf{55.93}
& 46.61{\scalebox{0.55}{$\pm$7.79}}
& 40.11{\scalebox{0.55}{$\pm$6.75}} \\

CHMNIST
& 29.06{\scalebox{0.55}{$\pm$0.00}}
& 49.53{\scalebox{0.55}{$\pm$1.90}}
& 45.48{\scalebox{0.55}{$\pm$2.53}}
& 30.57{\scalebox{0.55}{$\pm$6.17}}
& 33.60{\scalebox{0.55}{$\pm$3.78}}
& 34.06{\scalebox{0.55}{$\pm$9.48}}
& 50.82{\scalebox{0.55}{$\pm$5.69}}
& 50.931
& 12.50
& \textcolor{blue!70!black}{\underline{52.87{\scalebox{0.55}{$\pm$2.43}}}}
& \textbf{53.39{\scalebox{0.55}{$\pm$1.67}}} \\

COVID
& 45.84{\scalebox{0.55}{$\pm$0.00}}
& 44.69{\scalebox{0.55}{$\pm$1.03}}
& 43.52{\scalebox{0.55}{$\pm$11.06}}
& 42.35{\scalebox{0.55}{$\pm$8.13}}
& 42.30{\scalebox{0.55}{$\pm$8.20}}
& 37.50{\scalebox{0.55}{$\pm$2.70}}
& \textcolor{blue!70!black}{\underline{46.31{\scalebox{0.55}{$\pm$8.46}}}}
& 41.474
& 17.08
& 43.59{\scalebox{0.55}{$\pm$16.49}}
& \textbf{46.71{\scalebox{0.55}{$\pm$10.13}}} \\

LC25000
& 51.20{\scalebox{0.55}{$\pm$0.00}}
& 58.72{\scalebox{0.55}{$\pm$2.63}}
& 55.99{\scalebox{0.55}{$\pm$7.84}}
& 42.90{\scalebox{0.55}{$\pm$5.52}}
& 55.50{\scalebox{0.55}{$\pm$5.33}}
& 26.25{\scalebox{0.55}{$\pm$8.40}}
& \textcolor{blue!70!black}{\underline{64.97{\scalebox{0.55}{$\pm$4.24}}}}
& 37.840
& 20.00
& 61.88{\scalebox{0.55}{$\pm$2.99}}
& \textbf{68.61{\scalebox{0.55}{$\pm$4.67}}} \\

OCTMNIST
& 35.00{\scalebox{0.55}{$\pm$0.00}}
& \textbf{43.30{\scalebox{0.55}{$\pm$2.18}}}
& 32.43{\scalebox{0.55}{$\pm$2.00}}
& \textcolor{blue!70!black}{\underline{41.80{\scalebox{0.55}{$\pm$9.72}}}}
& 40.65{\scalebox{0.55}{$\pm$9.55}}
& 26.25{\scalebox{0.55}{$\pm$8.40}}
& 35.23{\scalebox{0.55}{$\pm$4.45}}
& 22.100
& 25.00
& 32.57{\scalebox{0.55}{$\pm$9.67}}
& 34.93{\scalebox{0.55}{$\pm$5.76}} \\

DermaMNIST
& \textbf{30.87{\scalebox{0.55}{$\pm$0.00}}}
& 19.63{\scalebox{0.55}{$\pm$3.27}}
& 16.99{\scalebox{0.55}{$\pm$1.25}}
& 21.20{\scalebox{0.55}{$\pm$15.79}}
& 18.20{\scalebox{0.55}{$\pm$7.34}}
& 14.66{\scalebox{0.55}{$\pm$3.09}}
& \textcolor{blue!70!black}{\underline{25.95{\scalebox{0.55}{$\pm$4.37}}}}
& 14.214
& 13.29
& 15.73{\scalebox{0.55}{$\pm$1.71}}
& 22.72{\scalebox{0.55}{$\pm$3.00}} \\

RetinaMNIST
& 27.68{\scalebox{0.55}{$\pm$0.00}}
& 33.81{\scalebox{0.55}{$\pm$5.17}}
& 39.36{\scalebox{0.55}{$\pm$6.29}}
& 28.43{\scalebox{0.55}{$\pm$1.81}}
& 30.60{\scalebox{0.55}{$\pm$4.05}}
& 28.75{\scalebox{0.55}{$\pm$4.53}}
& \textcolor{blue!70!black}{\underline{40.30{\scalebox{0.55}{$\pm$2.71}}}}
& 35.489
& 26.03
& 40.25{\scalebox{0.55}{$\pm$6.32}}
& \textbf{42.98{\scalebox{0.55}{$\pm$2.25}}} \\

Kvasir
& \textbf{55.50{\scalebox{0.55}{$\pm$0.00}}}
& 42.97{\scalebox{0.55}{$\pm$1.12}}
& 39.72{\scalebox{0.55}{$\pm$2.19}}
& 28.90{\scalebox{0.55}{$\pm$6.37}}
& 26.33{\scalebox{0.55}{$\pm$2.99}}
& 19.97{\scalebox{0.55}{$\pm$2.53}}
& 43.25{\scalebox{0.55}{$\pm$1.94}}
& 42.833
& 12.50
& 42.94{\scalebox{0.55}{$\pm$3.95}}
& \textcolor{blue!70!black}{\underline{46.80{\scalebox{0.55}{$\pm$2.79}}}} \\

CTKidney
& 36.73{\scalebox{0.55}{$\pm$0.00}}
& 31.96{\scalebox{0.55}{$\pm$4.00}}
& \textcolor{blue!70!black}{\underline{37.46{\scalebox{0.55}{$\pm$7.72}}}}
& \textbf{42.98{\scalebox{0.55}{$\pm$2.25}}}
& 32.19{\scalebox{0.55}{$\pm$3.82}}
& 27.56{\scalebox{0.55}{$\pm$13.87}}
& 34.30{\scalebox{0.55}{$\pm$9.48}}
& 27.127
& 29.80
& 32.75{\scalebox{0.55}{$\pm$3.17}}
& 29.03{\scalebox{0.55}{$\pm$6.35}} \\

KneeXray
& 27.48{\scalebox{0.55}{$\pm$0.00}}
& 31.62{\scalebox{0.55}{$\pm$5.73}}
& 34.42{\scalebox{0.55}{$\pm$2.92}}
& 24.83{\scalebox{0.55}{$\pm$7.42}}
& 30.25{\scalebox{0.55}{$\pm$9.78}}
& 19.10{\scalebox{0.55}{$\pm$6.73}}
& 26.59{\scalebox{0.55}{$\pm$1.75}}
& 25.543
& \textbf{38.59}
& \textcolor{blue!70!black}{\underline{38.30{\scalebox{0.55}{$\pm$1.18}}}}
& 26.67{\scalebox{0.55}{$\pm$2.30}} \\

\bottomrule
\end{tabular}
}
\end{table*}
Coming to the models, we use a Vision Mamba small variant (Vim-S) in the vision encoder and the transformer equivalent encoder at the text modality end. Further, the image resolution is resized to $224 \times 224$. Similar to the CoOp \cite{zhou2022conditional} model, \textbf{4} learnable context tokens are initialized. Importantly, the \underline{prompts} and \underline{Cross Mamba Interaction block} that consists of a combination of single and cross stacked mamba layers are kept trainable, and the remaining are kept frozen. The Stochastic Gradient Descent (SGD) is used with a momentum of \textbf{0.9} along with a cosine learning rate decay. A constant batch size of \textbf{16} is used across all the dataset and trained \textbf{100} epochs. The mean classification accuracy (top-1) is the primary metric reported here. As a comparison of performance and the complexity, we also report the model's GFLOP information.
Furthermore, we also conduct a set of ablation study with reference to the presence and the absence of the LLM model on the above mentioned datasets. All the experiments here are reported on the workstation powered with 2 NVIDIA Tesla V-100 with 32 GB of GPU.
\subsection{Comparison with Baselines}
\begin{figure}[h!]
    \centering
    \includegraphics[width=0.8\textwidth]{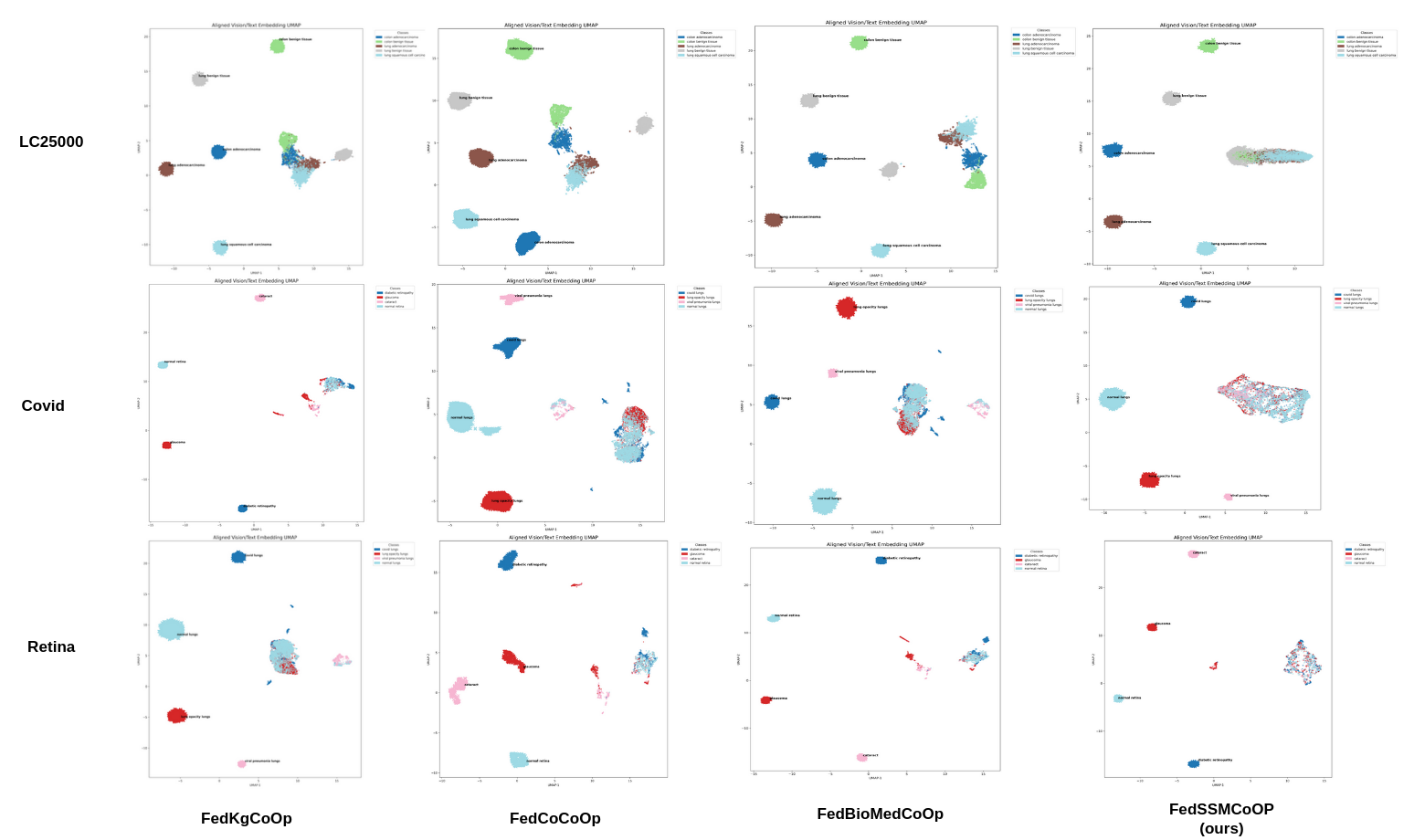}
    \caption{The UMAP plots of the proposed model against the baselines on different dataset. Observe that the proposed FedSSMCoOp framework has compact and well-separated embeddings in the text domain and a relatively looser cluster in the vision domain. This explains its ability of capturing discriminative features.}
    \label{fig:umap-analysis}
\end{figure}


The corresponding quantitative results of the baselines and the proposed works are shown in Table \ref{tab:main_results}. The results of the BiomedCLIP, CoCoOP, KgCoOp, and BiomedCoOp models, respectively, were not reported on the federated setting. Therefore, the above models were evaluated on the federated settings and further, this serves as the baseline. \\
It can observed with reference to the above models under the federated system, our proposed lightweight approach, in fact perform consistently in par or sometimes better than the baselines in some of the datasets. For instance, in the case of CHMNIST \cite{kather2016multi}, COVID, LC25000, there was approximate improvement in performance of roughly 4\%. On the other hand, in some instances, the proposed framework underperforms, particular in the case of BUSI and CTKidney datasets respectively. One of the reasons attributed to this could be due to the SSM's inability to capture fine-grained representations as compared to the attention-based framework which is effective in capturing this representation. In order to explain this, we use the UMAPs \cite{McInnes2018} and demonstrate the model's ability of capturing the features. The UMAPs of the proposed framework along with the baselines are shown in Figure \ref{fig:umap-analysis}. 

\begin{figure}[h!]
    \centering
    \includegraphics[width=0.8\textwidth]{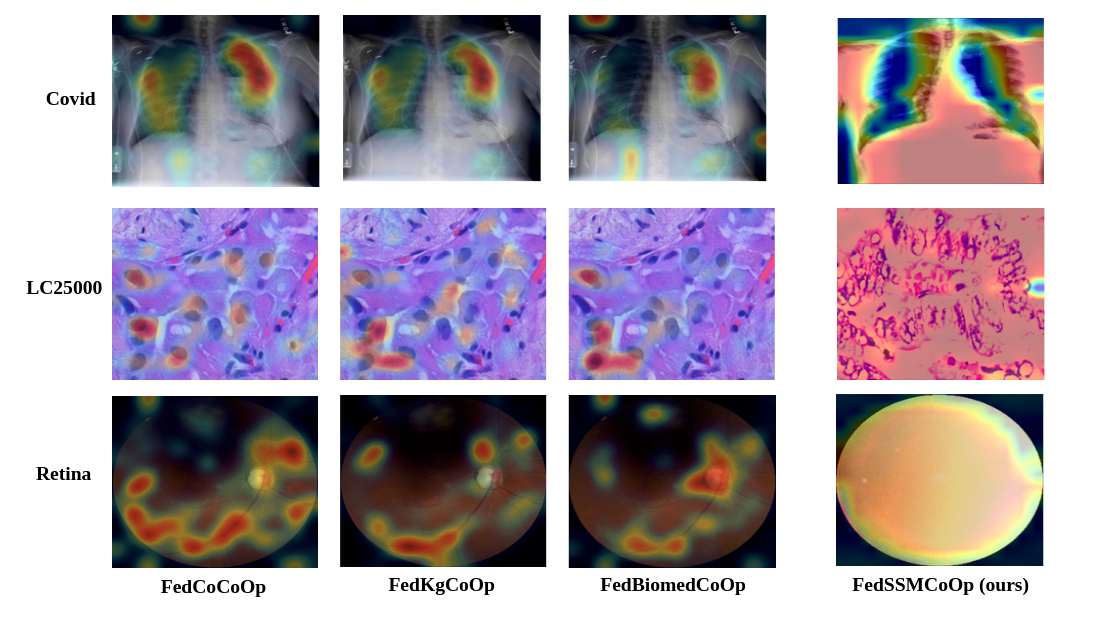}
    \caption{Visualization of activation maps of different models. The visualizations of baselines; FedCoCoOp, FedKgCoOp and FedBiomedCoOp are reported on 16 shot and the proposed FedSSMCoOp on 1 shot.}
    \label{fig:attention-analysis}
\end{figure}

Furthermore, in order to qualitative analysis the performance of the models, we plot the activation maps for the models under the three datasets, as shown in Figure \ref{fig:attention-analysis}.
\\ The models such as FedCoCoOP, FedKgCoOp and the FedBiomedCoOp respectively, perform good and concentrate on localized and fine features. For instance, in the case of the COVID dataset, most of the models particularly concentrate on the main part of the lungs. A similar scenario is observed in the case of the LC25000 dataset, where the model focuses more on the tissues. Importantly, in the Retina Dataset, the focus is not only around the cornea, but across the various other locations. Given the ability of these models to extract fine-grained features, these are ideal for detecting the abnormalities, especially in the terminal medical stages. \\
Furthermore, in the case of the FedSSMCoOp, a global perspective of attention is observed. Particularly in the case of the Covid dataset, the model focuses from the throat portion to almost the entire diaphragm section of the chest. But it is also to be noted, it doesn't focus on the localized portion. Similarly, in the case of the Retina dataset, the entire eyeball is attended by the model. This is vital especially in the scenario of cataract, where a global focus is necessary. Given this global focus ability of the model, this is ideal for an early-staged detection. On the other hand, a lower performance is observed on BUSI and CTKidney, where diagnosis relies on subtle lesion boundaries and highly localized pathological patterns. This behavior primarily reflects the global contextual modeling bias of the Vision Mamba backbone rather than the proposed federated prompt learning strategy. Despite this limitation, FedSSMCoOp provides an effective accuracy-efficiency trade-off across a diverse set of biomedical imaging modalities, reducing computational complexity by nearly $1.96\times$ while remaining competitive on most datasets.
\begin{table*}[h!]
\centering
\caption{Architectural and training-objective differences across the four ablation configurations. The LLM prototype bank $P_g$ is precomputed offline and never used during federated training; only the soft prompt is communicated.}
\label{tab:config_details}
\resizebox{\textwidth}{!}{%
\begin{tabular}{l|cccc}
\toprule
\textbf{Component / Objective} 
& \textbf{No-LLM (FedSSMCoOp)} 
& \textbf{LLM-Server} 
& \textbf{LLM-Client} 
& \textbf{LLM-Client+Server} \\
\midrule
LLM prototype bank $P_g$        & \textcolor{Maroon}{\ding{55}}         & Server-side       & Client-side       & Both \\
Client text branch              & Prompt only       & Prompt only       & Dual (prompt+$P_g$) & Dual (prompt+$P_g$) \\
Prompt--prototype fusion        & \textcolor{Maroon}{\ding{55}}         & \textcolor{Maroon}{\ding{55}}         & Learnable $\alpha$ & Learnable $\alpha$ \\
\midrule
Classification loss $\mathcal{L}_{CE}$        & \textcolor{DarkGreen}{\ding{51}} & \textcolor{DarkGreen}{\ding{51}} & \textcolor{DarkGreen}{\ding{51}} & \textcolor{DarkGreen}{\ding{51}} \\
Alignment loss $\mathcal{L}_{FedSCCM}$        & \textcolor{Maroon}{\ding{55}} & \textcolor{DarkGreen}{\ding{51}} & \textcolor{Maroon}{\ding{55}} & \textcolor{DarkGreen}{\ding{51}} \\
Distillation loss $\mathcal{L}_{FedKD}$       & \textcolor{Maroon}{\ding{55}} & \textcolor{Maroon}{\ding{55}} & \textcolor{Maroon}{\ding{55}} & \textcolor{DarkGreen}{\ding{51}} \\
\midrule
Total client objective 
& $\mathcal{L}_{CE}$ 
& $\mathcal{L}_{CE}{+}\lambda_{S}\mathcal{L}_{FedSCCM}$ 
& $\mathcal{L}_{CE}$ 
& $\mathcal{L}_{CE}{+}\lambda_{S}\mathcal{L}_{FedSCCM}{+}\lambda_{K}\mathcal{L}_{FedKD}$ \\
\midrule
Federated communication         & Prompt only       & Prompt only       & Prompt only       & Prompt only \\
LLM used during training         & \textcolor{Maroon}{\ding{55}}         & \textcolor{Maroon}{\ding{55}} (offline) & \textcolor{Maroon}{\ding{55}} (offline) & \textcolor{Maroon}{\ding{55}} (offline) \\
\bottomrule
\end{tabular}
}
\end{table*}
To investigate whether external semantic knowledge can further improve federated few-shot biomedical image classification, we conduct an ablation study by using a LLM into training pipeline. It is important to note that the proposed FedSSMCoOp framework is entirely \emph{LLM-free} during both training and inference. The LLM is introduced solely for this ablation study and is not part of the final proposed method. Specifically, prior to federated training, GPT-4 generates multiple textual descriptions for each disease class. These descriptions are then encoded using frozen BiomedCLIP text encoder, which is PubMedBERT-based, and averaged to obtain fixed-class semantic prototypes. Since these prototypes are computed only once before training, the LLM is never invoked during the communication round. Only the learnable prompt updates continue to be exchanged between the clients and the server. We evaluate on three strategies (Table~\ref{tab:config_details}): \textbf{LLM-Server}, where clients align prompt features to the global prototypes via $\mathcal{L}_{FedSCCM}=\text{MSE}(T,P_g)$; \textbf{LLM-Client}, where each client fuses prompt and prototype features through a learnable coefficient; and \textbf{LLM-Client+Server}, which combines both and adds a distillation loss $\mathcal{L}_{FedKD}=\text{KL}(\text{Pr}_{teacher}\|\text{Pr}_{student})$.

\begin{table*}[h!]
\centering
\caption{Ablation study on the proposed framework with inclusion of LLMs at the server and client ends across all datasets. For each of the datasets, \textbf{Best performing} and \textcolor{blue!70!black}{\underline{second best performing}} models are reported.}
\label{tab:ablation-llm-transpose}
\tiny
\resizebox{\textwidth}{!}{%
\begin{tabular}{l||c||ccc}
\toprule
\textbf{Dataset} & \textbf{No-LLM} & \textbf{LLM-Server} & \textbf{LLM-Client} & \textbf{LLM-Client+Server} \\
\midrule
\textbf{BTMRI} 
& \textbf{53.64{\scalebox{0.55}{$\pm$1.78}}} 
& 44.05{\scalebox{0.55}{$\pm$7.55}} 
& \textcolor{blue!70!black}{\underline{50.55{\scalebox{0.55}{$\pm$2.64}}}} 
& 46.11{\scalebox{0.55}{$\pm$6.81}} \\
\textbf{BUSI} 
& \textbf{40.11{\scalebox{0.55}{$\pm$6.75}}} 
& \textcolor{blue!70!black}{\underline{39.55{\scalebox{0.55}{$\pm$2.83}}}} 
& 39.41{\scalebox{0.55}{$\pm$9.32}} 
& 37.71{\scalebox{0.55}{$\pm$1.84}} \\
\textbf{CHMNIST} 
& \textcolor{blue!70!black}{\underline{53.39{\scalebox{0.55}{$\pm$1.67}}}} 
& 52.24{\scalebox{0.55}{$\pm$1.25}} 
& \textbf{54.01{\scalebox{0.55}{$\pm$2.25}}} 
& 52.22{\scalebox{0.55}{$\pm$1.47}} \\
\textbf{COVID} 
& \textbf{46.71{\scalebox{0.55}{$\pm$10.13}}} 
& 41.90{\scalebox{0.55}{$\pm$8.94}} 
& \textcolor{blue!70!black}{\underline{45.59{\scalebox{0.55}{$\pm$11.29}}}} 
& 41.99{\scalebox{0.55}{$\pm$11.27}} \\
\textbf{LC25000} 
& \textcolor{blue!70!black}{\underline{68.61{\scalebox{0.55}{$\pm$4.67}}}} 
& 64.38{\scalebox{0.55}{$\pm$6.39}} 
& \textbf{71.78{\scalebox{0.55}{$\pm$4.15}}} 
& 64.67{\scalebox{0.55}{$\pm$6.39}} \\
\textbf{OCTMNIST} 
& \textbf{34.93{\scalebox{0.55}{$\pm$5.76}}} 
& 31.13{\scalebox{0.55}{$\pm$5.92}} 
& \textcolor{blue!70!black}{\underline{32.73{\scalebox{0.55}{$\pm$6.22}}}} 
& 31.60{\scalebox{0.55}{$\pm$6.98}} \\
\textbf{DermaMNIST} 
& \textcolor{blue!70!black}{\underline{22.72{\scalebox{0.55}{$\pm$3.00}}}} 
& 17.06{\scalebox{0.55}{$\pm$5.39}} 
& \textbf{24.64{\scalebox{0.55}{$\pm$7.49}}} 
& 18.83{\scalebox{0.55}{$\pm$7.43}} \\
\textbf{RetinaMNIST} 
& \textcolor{blue!70!black}{\underline{42.98{\scalebox{0.55}{$\pm$2.25}}}} 
& \textbf{67.06{\scalebox{0.55}{$\pm$1.20}}} 
& 41.05{\scalebox{0.55}{$\pm$1.29}} 
& 39.83{\scalebox{0.55}{$\pm$6.16}} \\
\textbf{Kvasir} 
& \textcolor{blue!70!black}{\underline{46.80{\scalebox{0.55}{$\pm$2.79}}}} 
& \textbf{78.08{\scalebox{0.55}{$\pm$0.60}}} 
& 46.50{\scalebox{0.55}{$\pm$0.24}} 
& 41.61{\scalebox{0.55}{$\pm$2.06}} \\
\textbf{CTKidney} 
& 29.03{\scalebox{0.55}{$\pm$6.35}} 
& \textbf{29.36{\scalebox{0.55}{$\pm$2.37}}} 
& 27.80{\scalebox{0.55}{$\pm$3.79}} 
& \textcolor{blue!70!black}{\underline{29.04{\scalebox{0.55}{$\pm$1.47}}}} \\
\textbf{KneeXray} 
& \textbf{26.67{\scalebox{0.55}{$\pm$2.30}}} 
& 21.46{\scalebox{0.55}{$\pm$2.54}} 
& \textcolor{blue!70!black}{\underline{23.71{\scalebox{0.55}{$\pm$3.12}}}} 
& 22.12{\scalebox{0.55}{$\pm$5.57}} \\
\bottomrule
\end{tabular}
}
\end{table*}

Firstly, we conduct an ablation analysis depicting the impact of having on LLM either in server, client end or at both the ends. The results of the same are shown in Table \ref{tab:ablation-llm-transpose}. Despite the introduction of the LLM, there is only a minor improvement in the performance. It is worth noting, the performance is in-par and some cases better, when there is no LLM presence at the client and the server end respectively.
\begin{table*}[h!]
\centering
\caption{Ablation study (\% accuracy) across different model variants under 1-shot setting for all biomedical datasets. For each of the datasets, \textbf{Best performing} and \textcolor{blue!70!black}{\underline{second best performing}} models are reported.}
\label{tab:ablation-vit-vim-llm-transpose}
\resizebox{\textwidth}{!}{%
\begin{tabular}{lcccc||cc}
\toprule
\textbf{Dataset} & \textbf{ViT + no LLM} & \textbf{ViT + LLM} & \textbf{Vim + no LLM} & \textbf{Vim + LLM} & \textbf{Ours (w/o LLM)} & \textbf{Ours (with LLM)} \\
\midrule

\textbf{BTMRI}
& \textbf{58.33{\scalebox{0.55}{$\pm$11.79}}}
& 50.67{\scalebox{0.55}{$\pm$4.74}}
& 49.79{\scalebox{0.55}{$\pm$2.45}}
& 47.20{\scalebox{0.55}{$\pm$5.77}}
& \textcolor{blue!70!black}{\underline{53.64{\scalebox{0.55}{$\pm$1.78}}}}
& 44.05{\scalebox{0.55}{$\pm$7.55}} \\

\textbf{BUSI}
& \textcolor{blue!70!black}{\underline{41.24{\scalebox{0.55}{$\pm$2.04}}}}
& \textbf{46.61{\scalebox{0.55}{$\pm$7.79}}}
& 39.55{\scalebox{0.55}{$\pm$10.51}}
& 37.15{\scalebox{0.55}{$\pm$9.91}}
& 40.11{\scalebox{0.55}{$\pm$6.75}}
& 39.55{\scalebox{0.55}{$\pm$2.83}} \\

\textbf{CHMNIST}
& 42.75{\scalebox{0.55}{$\pm$2.38}}
& \textbf{57.87{\scalebox{0.55}{$\pm$2.43}}}
& 47.72{\scalebox{0.55}{$\pm$2.01}}
& 39.07{\scalebox{0.55}{$\pm$1.85}}
& \textcolor{blue!70!black}{\underline{53.39{\scalebox{0.55}{$\pm$1.67}}}}
& 52.24{\scalebox{0.55}{$\pm$1.25}} \\

\textbf{COVID}
& \textcolor{blue!70!black}{\underline{45.37{\scalebox{0.55}{$\pm$7.75}}}}
& 43.59{\scalebox{0.55}{$\pm$16.49}}
& 41.11{\scalebox{0.55}{$\pm$3.10}}
& 37.60{\scalebox{0.55}{$\pm$6.21}}
& \textbf{46.71{\scalebox{0.55}{$\pm$10.13}}}
& 41.90{\scalebox{0.55}{$\pm$8.94}} \\

\textbf{LC25000}
& 60.94{\scalebox{0.55}{$\pm$5.61}}
& 61.88{\scalebox{0.55}{$\pm$2.99}}
& \textcolor{blue!70!black}{\underline{65.49{\scalebox{0.55}{$\pm$2.25}}}}
& 51.48{\scalebox{0.55}{$\pm$2.78}}
& \textbf{68.61{\scalebox{0.55}{$\pm$4.67}}}
& 64.38{\scalebox{0.55}{$\pm$6.39}} \\

\textbf{OCTMNIST}
& 33.83{\scalebox{0.55}{$\pm$5.88}}
& 31.57{\scalebox{0.55}{$\pm$9.67}}
& \textbf{35.47{\scalebox{0.55}{$\pm$3.51}}}
& 28.90{\scalebox{0.55}{$\pm$4.53}}
& \textcolor{blue!70!black}{\underline{34.93{\scalebox{0.55}{$\pm$5.76}}}}
& 31.13{\scalebox{0.55}{$\pm$5.92}} \\

\textbf{DermaMNIST}
& 17.12{\scalebox{0.55}{$\pm$4.78}}
& 15.73{\scalebox{0.55}{$\pm$1.71}}
& \textcolor{blue!70!black}{\underline{22.94{\scalebox{0.55}{$\pm$8.17}}}}
& \textbf{25.23{\scalebox{0.55}{$\pm$6.47}}}
& 22.72{\scalebox{0.55}{$\pm$3.00}}
& 17.06{\scalebox{0.55}{$\pm$5.39}} \\

\textbf{RetinaMNIST}
& \textcolor{blue!70!black}{\underline{45.56{\scalebox{0.55}{$\pm$2.24}}}}
& 40.25{\scalebox{0.55}{$\pm$6.32}}
& 39.20{\scalebox{0.55}{$\pm$3.83}}
& 38.03{\scalebox{0.55}{$\pm$2.96}}
& 42.98{\scalebox{0.55}{$\pm$2.25}}
& \textbf{67.06{\scalebox{0.55}{$\pm$1.20}}} \\

\textbf{Kvasir}
& 39.47{\scalebox{0.55}{$\pm$3.14}}
& 42.94{\scalebox{0.55}{$\pm$3.95}}
& 39.41{\scalebox{0.55}{$\pm$4.62}}
& 38.03{\scalebox{0.55}{$\pm$2.96}}
& \textcolor{blue!70!black}{\underline{46.80{\scalebox{0.55}{$\pm$2.79}}}}
& \textbf{78.08{\scalebox{0.55}{$\pm$0.60}}} \\

\textbf{CTKidney}
& \textcolor{blue!70!black}{\underline{30.83{\scalebox{0.55}{$\pm$5.45}}}}
& \textbf{32.75{\scalebox{0.55}{$\pm$3.17}}}
& 25.70{\scalebox{0.55}{$\pm$2.75}}
& 28.79{\scalebox{0.55}{$\pm$3.09}}
& 29.03{\scalebox{0.55}{$\pm$6.35}}
& 29.36{\scalebox{0.55}{$\pm$2.37}} \\

\textbf{KneeXray}
& 24.40{\scalebox{0.55}{$\pm$2.38}}
& \textbf{38.30{\scalebox{0.55}{$\pm$1.18}}}
& \textcolor{blue!70!black}{\underline{27.42{\scalebox{0.55}{$\pm$3.42}}}}
& 24.16{\scalebox{0.55}{$\pm$3.62}}
& 26.67{\scalebox{0.55}{$\pm$2.30}}
& 21.46{\scalebox{0.55}{$\pm$2.54}} \\

\bottomrule
\end{tabular}
}
\end{table*}
We also conduct an ablation study reported in Table \ref{tab:ablation-vit-vim-llm-transpose}, with the different combinations of either attention mechanism of ViT and the SSM mechanism of Vision Mamba (Vim) integrated with the LLM. This is then also compared with the proposed framework of FedSSMCoOp. In particular the Small variant of Vision Mamba (Vim-S) is chosen and the Base variant of Vision Transformer (ViT-B) was compared as used by \cite{koleilat2025biomedcoop}. Importantly the Gpt4 \cite{achiam2023gpt} is used as the LLM here, and is integrated at the server side of the federated setup. \\
One of the main observations is that, the propose framework with Vision mamba and COMO based cross-modal interaction without the use of LLM in the server performs consistently in-par or sometimes better than the baselines. For instance, there is a good improvement observed in the case of the COVID, LC25000, and Kvasir datasets. Further with RetinaMNIST, OCTMNIST and CHMNIST datasets, despite the presence of LLM helped improving the accuracy, the proposed approach was also within 5\% range of the prior approach.
\subsection{Effect of Client Participation}
In the further extension, another ablation study was performed on the proposed FedSSMCoOp, with the effect of the number of clients under three Dirichlet values ($\alpha$) 0.1, 0.5 and 1 respectively, thereby giving an indication on how the model performs under the non-IID scenarios. The results on the corresponding datasets are reported in Table \ref{tab:client_participation}. It can be observed that the model performance gap between the lowest Dirichlet value as compared to the highest Dirichlet value is quite less. This indicates that the model on average is relative stable under non-IID conditions when analysed across most of the datasets. \\
\begin{table*}[h!]
\centering
\caption{Effect of client count and data heterogeneity on FedSSMCoOp accuracy (\%) at 1-shot, averaged over 3 seeds. For each of the datasets, \textbf{Best performing} and \textcolor{blue!70!black}{\underline{second best performing}} models are reported.}
\label{tab:client_participation}
\renewcommand{\arraystretch}{1.2}
\setlength{\tabcolsep}{3pt}
\scriptsize
\resizebox{\textwidth}{!}{%
\begin{tabular}{l|ccc|ccc|ccc}
\toprule
\textbf{Dataset} 
& \multicolumn{3}{c|}{\textbf{4 Clients}} 
& \multicolumn{3}{c|}{\textbf{8 Clients}} 
& \multicolumn{3}{c}{\textbf{16 Clients}} \\
\cmidrule(lr){2-4} \cmidrule(lr){5-7} \cmidrule(lr){8-10}
& $\alpha$=0.1 & $\alpha$=0.5 & $\alpha$=1.0 
& $\alpha$=0.1 & $\alpha$=0.5 & $\alpha$=1.0 
& $\alpha$=0.1 & $\alpha$=0.5 & $\alpha$=1.0 \\
\midrule

BTMRI
& 31.59{\scalebox{0.55}{$\pm$12.96}}
& 32.17{\scalebox{0.55}{$\pm$12.90}}
& 32.81{\scalebox{0.55}{$\pm$15.43}}
& 33.72{\scalebox{0.55}{$\pm$15.92}}
& 53.64{\scalebox{0.55}{$\pm$1.78}}
& \textcolor{blue!70!black}{\underline{42.46{\scalebox{0.55}{$\pm$6.20}}}}
& 34.50{\scalebox{0.55}{$\pm$15.50}}
& \textbf{52.21{\scalebox{0.55}{$\pm$7.94}}}
& 33.59{\scalebox{0.55}{$\pm$16.13}} \\

BUSI
& 40.53{\scalebox{0.55}{$\pm$10.81}}
& 40.96{\scalebox{0.55}{$\pm$9.76}}
& 41.24{\scalebox{0.55}{$\pm$10.81}}
& 40.82{\scalebox{0.55}{$\pm$11.71}}
& 40.11{\scalebox{0.55}{$\pm$6.75}}
& 40.11{\scalebox{0.55}{$\pm$10.17}}
& 41.00{\scalebox{0.55}{$\pm$11.20}}
& \textcolor{blue!70!black}{\underline{42.30{\scalebox{0.55}{$\pm$11.00}}}}
& 38.72{\scalebox{0.55}{$\pm$5.63}} \\

CHMNIST
& 36.37{\scalebox{0.55}{$\pm$3.80}}
& 36.30{\scalebox{0.55}{$\pm$4.84}}
& 36.19{\scalebox{0.55}{$\pm$4.37}}
& 33.38{\scalebox{0.55}{$\pm$4.91}}
& \textbf{53.39{\scalebox{0.55}{$\pm$1.67}}}
& \textcolor{blue!70!black}{\underline{36.39{\scalebox{0.55}{$\pm$3.82}}}}
& 34.22{\scalebox{0.55}{$\pm$4.82}}
& 35.18{\scalebox{0.55}{$\pm$4.15}}
& \textbf{37.20{\scalebox{0.55}{$\pm$5.94}}} \\

COVID
& 34.84{\scalebox{0.55}{$\pm$6.65}}
& 34.82{\scalebox{0.55}{$\pm$6.49}}
& 34.56{\scalebox{0.55}{$\pm$7.49}}
& 34.10{\scalebox{0.55}{$\pm$7.20}}
& \textbf{46.71{\scalebox{0.55}{$\pm$10.13}}}
& \textcolor{blue!70!black}{\underline{34.88{\scalebox{0.55}{$\pm$6.62}}}}
& 33.68{\scalebox{0.55}{$\pm$6.99}}
& 34.29{\scalebox{0.55}{$\pm$6.58}}
& \textbf{34.84{\scalebox{0.55}{$\pm$7.12}}} \\

LC25000
& 48.56{\scalebox{0.55}{$\pm$3.63}}
& 48.77{\scalebox{0.55}{$\pm$3.72}}
& \textcolor{blue!70!black}{\underline{49.17{\scalebox{0.55}{$\pm$3.41}}}}
& 48.49{\scalebox{0.55}{$\pm$2.79}}
& \textbf{68.61{\scalebox{0.55}{$\pm$4.67}}}
& 48.67{\scalebox{0.55}{$\pm$3.41}}
& 48.54{\scalebox{0.55}{$\pm$3.07}}
& 48.72{\scalebox{0.55}{$\pm$3.31}}
& \textbf{48.92{\scalebox{0.55}{$\pm$2.92}}} \\

OCTMNIST
& 25.70{\scalebox{0.55}{$\pm$6.17}}
& 25.87{\scalebox{0.55}{$\pm$5.72}}
& \textcolor{blue!70!black}{\underline{26.07{\scalebox{0.55}{$\pm$5.10}}}}
& 25.27{\scalebox{0.55}{$\pm$5.34}}
& 34.93{\scalebox{0.55}{$\pm$5.76}}
& \textbf{26.20{\scalebox{0.55}{$\pm$5.45}}}
& 25.27{\scalebox{0.55}{$\pm$5.85}}
& 24.90{\scalebox{0.55}{$\pm$1.44}}
& 25.60{\scalebox{0.55}{$\pm$5.37}} \\

DermaMNIST
& 21.73{\scalebox{0.55}{$\pm$16.37}}
& 22.01{\scalebox{0.55}{$\pm$16.43}}
& \textcolor{blue!70!black}{\underline{22.43{\scalebox{0.55}{$\pm$16.29}}}}
& 22.44{\scalebox{0.55}{$\pm$17.01}}
& \textcolor{blue!70!black}{\underline{22.72{\scalebox{0.55}{$\pm$3.00}}}}
& 22.43{\scalebox{0.55}{$\pm$16.33}}
& 21.80{\scalebox{0.55}{$\pm$16.50}}
& \textbf{45.65{\scalebox{0.55}{$\pm$4.72}}}
& 21.10{\scalebox{0.55}{$\pm$16.97}} \\

RetinaMNIST
& 26.03{\scalebox{0.55}{$\pm$1.89}}
& \textcolor{blue!70!black}{\underline{26.42{\scalebox{0.55}{$\pm$1.70}}}}
& 26.79{\scalebox{0.55}{$\pm$2.06}}
& 26.42{\scalebox{0.55}{$\pm$1.79}}
& \textbf{42.98{\scalebox{0.55}{$\pm$2.25}}}
& 26.34{\scalebox{0.55}{$\pm$1.36}}
& 26.31{\scalebox{0.55}{$\pm$1.73}}
& \textbf{29.60{\scalebox{0.55}{$\pm$1.25}}}
& 26.42{\scalebox{0.55}{$\pm$1.79}} \\

Kvasir
& 38.31{\scalebox{0.55}{$\pm$6.30}}
& 38.59{\scalebox{0.55}{$\pm$5.13}}
& 38.67{\scalebox{0.55}{$\pm$5.55}}
& \textbf{48.77{\scalebox{0.55}{$\pm$3.72}}}
& 46.80{\scalebox{0.55}{$\pm$2.79}}
& 38.09{\scalebox{0.55}{$\pm$4.88}}
& 38.62{\scalebox{0.55}{$\pm$5.24}}
& \textcolor{blue!70!black}{\underline{39.42{\scalebox{0.55}{$\pm$4.19}}}}
& 37.50{\scalebox{0.55}{$\pm$5.35}} \\

CTKidney
& 39.05{\scalebox{0.55}{$\pm$5.95}}
& \textbf{43.77{\scalebox{0.55}{$\pm$2.55}}}
& 39.53{\scalebox{0.55}{$\pm$2.51}}
& 38.34{\scalebox{0.55}{$\pm$5.12}}
& 29.03{\scalebox{0.55}{$\pm$6.35}}
& 38.65{\scalebox{0.55}{$\pm$5.55}}
& 39.14{\scalebox{0.55}{$\pm$5.02}}
& \textcolor{blue!70!black}{\underline{43.31{\scalebox{0.55}{$\pm$3.82}}}}
& 38.40{\scalebox{0.55}{$\pm$5.15}} \\

KneeXray
& 19.66{\scalebox{0.55}{$\pm$5.62}}
& 19.46{\scalebox{0.55}{$\pm$4.91}}
& \textbf{20.61{\scalebox{0.55}{$\pm$5.16}}}
& 18.98{\scalebox{0.55}{$\pm$6.00}}
& 26.67{\scalebox{0.55}{$\pm$2.30}}
& \textcolor{blue!70!black}{\underline{20.07{\scalebox{0.55}{$\pm$6.57}}}}
& 19.37{\scalebox{0.55}{$\pm$5.98}}
& 20.53{\scalebox{0.55}{$\pm$5.66}}
& 18.82{\scalebox{0.55}{$\pm$6.40}} \\

\bottomrule
\end{tabular}
}
\end{table*}

In particular the highest stability was observed with the 4-client system, where the deviation across different $\alpha$ values was less than 2\%. A similar scenario is observed in the case of the reported 8-client system, with a minor spike in the gap especially in the case of BTMRI and KVASIR datasets respectively. Generalizing it to a compute intensive 16-client system, a similar pattern followed up, with minimal fluctuations in the gap. In addition, the ablation corresponding to different shots from 2, 4, 8 and 16 respectively is performed and the performances are reported in the Appendix. 
\subsection{Efficiency Analysis}
We then made an analysis of the computation complexity in terms of GFLOPs for each of the models as shown in Table \ref{tab:efficiency-all}. Given the different training strategies amongst the respective datasets, a variation can be observed here. 
\begin{table*}[h!]
\centering
\caption{The corresponding computational complexity (GFLOPs) with respect to the baselines and the proposed framework across all biomedical datasets.}
\label{tab:efficiency-all}
\renewcommand{\arraystretch}{1.1}
\setlength{\tabcolsep}{6pt}

\resizebox{\textwidth}{!}{%
\begin{tabular}{lccccccccccc}
\toprule
\textbf{Model} 
& \textbf{BTMRI} 
& \textbf{BUSI} 
& \textbf{CHMNIST} 
& \textbf{COVID} 
& \textbf{LC25000} 
& \textbf{OCTMNIST}
& \textbf{DermaMNIST}
& \textbf{RetinaMNIST}
& \textbf{Kvasir}
& \textbf{CTKidney}
& \textbf{KneeXray} \\
\midrule

\textbf{FedBiomedCLIP}
& 33.70G
& 33.70G
& 33.70G
& 33.70G
& 33.70G
& 33.70G
& 33.70G
& 33.70G
& 33.70G
& 33.70G
& 33.70G \\

\textbf{FedCoCoOp}
& 38.04G
& 34.17G
& 53.55G
& 38.04G
& 38.04G
& 49.67G 
& 41.92G 
& 53.55G 
& 41.92G
& 38.04G
& 38.04G \\

\textbf{FedKgCoOp}
& 38.04G
& 34.17G
& 53.55G
& 38.04G
& 38.04G
& 49.67G 
& 41.61G 
& 53.55G 
& 41.92G
& 38.04G
& 38.04G \\

\textbf{PromptFL}  
& 26.33G 
& 22.46G 
& 41.84G 
& 26.33G 
& 30.21G 
& 26.33G
& 37.96G
& 26.33G
& 41.84G
& 26.33G
& 30.21G \\

\textbf{PromptFL + FedProx}  
& 26.33G 
& 22.46G 
& 41.84G 
& 26.33G 
& 30.21G 
& 26.33G
& 37.96G
& 26.33G
& 41.84G
& 26.33G
& 30.21G \\

\textbf{FedOTP}  
& 41.94G 
& 34.13G 
& 73.14G 
& 41.94G 
& 49.74G 
& 41.94G
& 65.34G
& 49.74G
& 73.14G
& 41.94G
& 41.94G \\

\textbf{FedBiomedCoOp}  
& 38.04G 
& 34.16G 
& 53.54G 
& 38.04G 
& 41.91G 
& 38.04G
& 49.66G
& 38.04G
& 53.54G
& 38.04G
& 41.91G \\ \hline

\textbf{FedSSMCoOp (Ours)} 
& \textbf{15.81G}
& \textbf{11.95G} 
& \textbf{31.33G} 
& \textbf{15.81G}
& \textbf{15.81G} 
& \textbf{27.45G} 
& \textbf{19.69G}
& \textbf{31.32G}
& \textbf{19.69G}
& \textbf{15.81G}
& \textbf{15.81G} \\ \hline
\end{tabular}
}
\end{table*}
The FedBioMedCLIP and the FedBioMedCoOp model make use of the BERT and the ViT-B for the text and image encodings respectively, which make them heavy from the computational perspective. The FedCoCop and the FedKgCoOp introduce the concept of learnable prompts in addition to the ViT based vision encoderand transformer equivalent text encoder respectively. Despite being heavy, these are relatively less bulky than the prior models. On the other hand, the PromptFL and FedOTP variant frameworks exploit ResNet based vision encoder additionally with learnable prompts, thus making the system relatively light. \\
However, our proposed FedSSMCoop makes use of vision mamba for the vision encoding with an introduction of a learnable Cross Mamba block for better cross modal interaction. The computational complexity (GFLOPs) reported in this work was measured as the number of floating-point operations required for a single forward pass using an input image of size $224\times224$. The reported values include the inference pipeline consisting of the frozen visual encoder, text encoder, prompt-learning modules, and the proposed Mamba Interaction Block (MIB). Despite an increase in the total parameter count, the overall GFLOPs are actually quite lower than most of the baselines used above. This makes it an apt candidate for edge-device deployment.
\section{Conclusion \& Future Directions}
We introduce FedSSMCoOp, an SSM-based federated few-shot framework for biomedical image classification that uses Vision Mamba and the Mamba interaction block for cross-modal interaction. This helps in capturing the cross-modal information between the two modalities, unlike the prior works, which usually rely on the cosine similarity between the pooled modality features. Since, the mamba interaction block already captures the cross-modal information and alignments of the modality features, the LLM presence, which performed the same task can be removed. Further, by communicating with just the soft prompt deltas between clients and server, the model is optimized for computation and performance. The FedSSMCoOp is trained and analyzed on different biomedical datasets, and the performances are quite consistent and stable against the baselines, given that the proposed approach is roughly \textbf{1.96} times lighter than other models. From the UMAPs and the activation maps, the FedSSMCoOp is a suitable model for early detection and at the inference stage. Despite the positives, there are a few drawbacks and future directions that can be carried out. \\
The Mamba-based approach fails to capture the fine-grained representations, unlike the attention-based approach. Despite the high computation factor, the attention-based approaches are still dominant even today. Thus, the future directions can look at a hybrid approach of Attention and State space mechanism \cite{hatamizadeh2025mambavision}. \\
Secondly, there were instances where the LLM presence in the server side of the federated learning aided in the performance especially in Kvasir and RetinaMNIST dataset due to their presence of localized detections. The forthcoming works can also focus more on whether to what extent the LLM help to aid these alignments.

\bibliography{main}

@article{guo2023promptfl,
  title={Promptfl: Let federated participants cooperatively learn prompts instead of models--federated learning in age of foundation model},
  author={Guo, Tao and Guo, Song and Wang, Junxiao and Tang, Xueyang and Xu, Wenchao},
  journal={IEEE Transactions on Mobile Computing},
  volume={23},
  number={5},
  pages={5179--5194},
  year={2023},
  publisher={IEEE}
}

@inproceedings{mcmahan2017communication,
  title={Communication-efficient learning of deep networks from decentralized data},
  author={McMahan, Brendan and Moore, Eider and Ramage, Daniel and Hampson, Seth and y Arcas, Blaise Aguera},
  booktitle={Artificial intelligence and statistics},
  pages={1273--1282},
  year={2017},
  organization={PMLR}
}

@article{li2020federated,
  title={Federated optimization in heterogeneous networks},
  author={Li, Tian and Sahu, Anit Kumar and Zaheer, Manzil and Sanjabi, Maziar and Talwalkar, Ameet and Smith, Virginia},
  journal={Proceedings of Machine learning and systems},
  volume={2},
  pages={429--450},
  year={2020}
}

@inproceedings{karimireddy2020scaffold,
  title={Scaffold: Stochastic controlled averaging for federated learning},
  author={Karimireddy, Sai Praneeth and Kale, Satyen and Mohri, Mehryar and Reddi, Sashank and Stich, Sebastian and Suresh, Ananda Theertha},
  booktitle={International conference on machine learning},
  pages={5132--5143},
  year={2020},
  organization={PMLR}
}

@article{zhou2022learning,
  title={Learning to prompt for vision-language models},
  author={Zhou, Kaiyang and Yang, Jingkang and Loy, Chen Change and Liu, Ziwei},
  journal={International journal of computer vision},
  volume={130},
  number={9},
  pages={2337--2348},
  year={2022},
  publisher={Springer}
}

@inproceedings{zhou2022conditional,
  title={Conditional prompt learning for vision-language models},
  author={Zhou, Kaiyang and Yang, Jingkang and Loy, Chen Change and Liu, Ziwei},
  booktitle={Proceedings of the IEEE/CVF conference on computer vision and pattern recognition},
  pages={16816--16825},
  year={2022}
}

@inproceedings{yao2023visual,
  title={Visual-language prompt tuning with knowledge-guided context optimization},
  author={Yao, Hantao and Zhang, Rui and Xu, Changsheng},
  booktitle={Proceedings of the IEEE/CVF conference on computer vision and pattern recognition},
  pages={6757--6767},
  year={2023}
}

@inproceedings{lu2022prompt,
  title={Prompt distribution learning},
  author={Lu, Yuning and Liu, Jianzhuang and Zhang, Yonggang and Liu, Yajing and Tian, Xinmei},
  booktitle={Proceedings of the IEEE/CVF conference on computer vision and pattern recognition},
  pages={5206--5215},
  year={2022}
}

@inproceedings{koleilat2025biomedcoop,
  title={Biomedcoop: Learning to prompt for biomedical vision-language models},
  author={Koleilat, Taha and Asgariandehkordi, Hojat and Rivaz, Hassan and Xiao, Yiming},
  booktitle={Proceedings of the Computer Vision and Pattern Recognition Conference},
  pages={14766--14776},
  year={2025}
}

@article{gu2023mamba,
  title={Mamba: Linear-time sequence modeling with selective state spaces},
  author={Gu, Albert and Dao, Tri},
  journal={arXiv preprint arXiv:2312.00752},
  year={2023}
}

@article{zhu2024vision,
  title={Vision mamba: Efficient visual representation learning with bidirectional state space model},
  author={Zhu, Lianghui and Liao, Bencheng and Zhang, Qian and Wang, Xinlong and Liu, Wenyu and Wang, Xinggang},
  journal={arXiv preprint arXiv:2401.09417},
  year={2024}
}

@article{liu2024vmamba,
  title={Vmamba: Visual state space model},
  author={Liu, Yue and Tian, Yunjie and Zhao, Yuzhong and Yu, Hongtian and Xie, Lingxi and Wang, Yaowei and Ye, Qixiang and Jiao, Jianbin and Liu, Yunfan},
  journal={Advances in neural information processing systems},
  volume={37},
  pages={103031--103063},
  year={2024}
}

@article{yue2024medmamba,
  title={Medmamba: Vision mamba for medical image classification},
  author={Yue, Yubiao and Li, Zhenzhang},
  journal={arXiv preprint arXiv:2403.03849},
  year={2024}
}

@article{li2024mambadfuse,
  title={Mambadfuse: A mamba-based dual-phase model for multi-modality image fusion},
  author={Li, Zhe and Pan, Haiwei and Zhang, Kejia and Wang, Yuhua and Yu, Fengming},
  journal={arXiv preprint arXiv:2404.08406},
  year={2024}
}

@article{liu2026cross,
  title={COMO: Cross-mamba interaction and offset-guided fusion for multimodal object detection},
  author={Liu, Chang and Ma, Xin and Yang, Xiaochen and Zhang, Yuxiang and Dong, Yanni},
  journal={Information Fusion},
  volume={125},
  pages={103414},
  year={2026},
  publisher={Elsevier}
}

@inproceedings{radford2021learning,
  title={Learning transferable visual models from natural language supervision},
  author={Radford, Alec and Kim, Jong Wook and Hallacy, Chris and Ramesh, Aditya and Goh, Gabriel and Agarwal, Sandhini and Sastry, Girish and Askell, Amanda and Mishkin, Pamela and Clark, Jack and others},
  booktitle={International conference on machine learning},
  pages={8748--8763},
  year={2021},
  organization={PmLR}
}

@article{kather2016multi,
  title={Multi-class texture analysis in colorectal cancer histology Sci},
  author={Kather, Jakob Nikolas and Weis, Cleo-Aron and Bianconi, Francesco and Melchers, Susanne M and Schad, Lothar R and Gaiser, Timo and Marx, Alexander and Z{\"o}llner, Frank Gerrit},
  journal={Rep},
  volume={6},
  number={1},
  pages={1--11},
  year={2016}
}

@inproceedings{guo2023pfedprompt,
  title={Pfedprompt: Learning personalized prompt for vision-language models in federated learning},
  author={Guo, Tao and Guo, Song and Wang, Junxiao},
  booktitle={Proceedings of the ACM Web Conference 2023},
  pages={1364--1374},
  year={2023}
}

@article{kermany2018identifying,
  title={Identifying medical diagnoses and treatable diseases by image-based deep learning},
  author={Kermany, Daniel S and Goldbaum, Michael and Cai, Wenjia and Valentim, Carolina CS and Liang, Huiying and Baxter, Sally L and McKeown, Alex and Yang, Ge and Wu, Xiaokang and Yan, Fangbing and others},
  journal={cell},
  volume={172},
  number={5},
  pages={1122--1131},
  year={2018},
  publisher={Elsevier}
}

@article{al2020dataset,
  title={Dataset of breast ultrasound images},
  author={Al-Dhabyani, Walid and Gomaa, Mohammed and Khaled, Hussien and Fahmy, Aly},
  journal={Data in brief},
  volume={28},
  pages={104863},
  year={2020},
  publisher={Elsevier}
}

@article{borkowski1912lung,
  title={Lung and colon cancer histopathological image dataset (lc25000)(2019)},
  author={Borkowski, Andrew A and Bui, Marilyn M and Thomas, L Brannon and Wilson, Catherine P and DeLand, Lauren A and Mastorides, Stephen M},
  journal={arXiv preprint arXiv:1912.12142},
  year={1912}
}

@article{tahir2021covid,
  title={COVID-19 infection localization and severity grading from chest X-ray images},
  author={Tahir, Anas M and Chowdhury, Muhammad EH and Khandakar, Amith and Rahman, Tawsifur and Qiblawey, Yazan and Khurshid, Uzair and Kiranyaz, Serkan and Ibtehaz, Nabil and Rahman, M Sohel and Al-Maadeed, Somaya and others},
  journal={Computers in biology and medicine},
  volume={139},
  pages={105002},
  year={2021},
  publisher={Elsevier}
}

@article{oord2018representation,
  title={Representation learning with contrastive predictive coding},
  author={Oord, Aaron van den and Li, Yazhe and Vinyals, Oriol},
  journal={arXiv preprint arXiv:1807.03748},
  year={2018}
}

@inproceedings{zhai2023sigmoid,
  title={Sigmoid loss for language image pre-training},
  author={Zhai, Xiaohua and Mustafa, Basil and Kolesnikov, Alexander and Beyer, Lucas},
  booktitle={Proceedings of the IEEE/CVF international conference on computer vision},
  pages={11975--11986},
  year={2023}
}

@inproceedings{li2022blip,
  title={Blip: Bootstrapping language-image pre-training for unified vision-language understanding and generation},
  author={Li, Junnan and Li, Dongxu and Xiong, Caiming and Hoi, Steven},
  booktitle={International conference on machine learning},
  pages={12888--12900},
  year={2022},
  organization={PMLR}
}

@article{achiam2023gpt,
  title={Gpt-4 technical report},
  author={Achiam, Josh and Adler, Steven and Agarwal, Sandhini and Ahmad, Lama and Akkaya, Ilge and Aleman, Florencia Leoni and Almeida, Diogo and Altenschmidt, Janko and Altman, Sam and Anadkat, Shyamal and others},
  journal={arXiv preprint arXiv:2303.08774},
  year={2023}
}

@article{chen2018knee,
  title={Knee osteoarthritis severity grading dataset},
  author={Chen, Pingjun},
  journal={Mendeley Data},
  volume={1},
  number={10.17632},
  pages={30784984},
  year={2018}
}

@inproceedings{pogorelov2017kvasir,
  title={Kvasir: A multi-class image dataset for computer aided gastrointestinal disease detection},
  author={Pogorelov, Konstantin and Randel, Kristin Ranheim and Griwodz, Carsten and Eskeland, Sigrun Losada and de Lange, Thomas and Johansen, Dag and Spampinato, Concetto and Dang-Nguyen, Duc-Tien and Lux, Mathias and Schmidt, Peter Thelin and others},
  booktitle={Proceedings of the 8th ACM on Multimedia Systems Conference},
  pages={164--169},
  year={2017}
}

@article{codella2019skin,
  title={Skin lesion analysis toward melanoma detection 2018: A challenge hosted by the international skin imaging collaboration (isic)},
  author={Codella, Noel and Rotemberg, Veronica and Tschandl, Philipp and Celebi, M Emre and Dusza, Stephen and Gutman, David and Helba, Brian and Kalloo, Aadi and Liopyris, Konstantinos and Marchetti, Michael and others},
  journal={arXiv preprint arXiv:1902.03368},
  year={2019}
}

@article{tschandl2018ham10000,
  title={The HAM10000 dataset, a large collection of multi-source dermatoscopic images of common pigmented skin lesions},
  author={Tschandl, Philipp and Rosendahl, Cliff and Kittler, Harald},
  journal={Scientific data},
  volume={5},
  number={1},
  pages={180161},
  year={2018},
  publisher={Nature Publishing Group}
}

@inproceedings{kohler2013automatic,
  title={Automatic no-reference quality assessment for retinal fundus images using vessel segmentation},
  author={K{\"o}hler, Thomas and Budai, Attila and Kraus, Martin F and Odstr{\v{c}}ilik, Jan and Michelson, Georg and Hornegger, Joachim},
  booktitle={Proceedings of the 26th IEEE international symposium on computer-based medical systems},
  pages={95--100},
  year={2013},
  organization={IEEE}
}

@article{islam2022vision,
  title={Vision transformer and explainable transfer learning models for auto detection of kidney cyst, stone and tumor from CT-radiography},
  author={Islam, Md Nazmul and Hasan, Mehedi and Hossain, Md Kabir and Alam, Md Golam Rabiul and Uddin, Md Zia and Soylu, Ahmet},
  journal={Scientific Reports},
  volume={12},
  number={1},
  pages={11440},
  year={2022},
  publisher={Nature Publishing Group UK London}
}

@article{McInnes2018,
  title = {UMAP: Uniform Manifold Approximation and Projection},
  author = {McInnes, Leland and Healy, John and Melville, James},
  journal = {Journal of Open Source Software},
  volume = {3},
  number = {29},
  pages = {861},
  year = {2018},
  doi = {10.21105/joss.00861},
  url = {https://doi.org}
}

@inproceedings{tan-bansal-2019-lxmert,
    title = "{LXMERT}: Learning Cross-Modality Encoder Representations from Transformers",
    author = "Tan, Hao and Bansal, Mohit",
    booktitle = "Proceedings of the 2019 Conference on Empirical Methods in Natural Language Processing and the 9th International Joint Conference on Natural Language Processing (EMNLP-IJCNLP)",
    year = "2019",
    pages = "5100--5111",
    url = "https://aclanthology.org/D19-1514"
}

@inproceedings{hatamizadeh2025mambavision,
  title={Mambavision: A hybrid mamba-transformer vision backbone},
  author={Hatamizadeh, Ali and Kautz, Jan},
  booktitle={Proceedings of the Computer Vision and Pattern Recognition Conference},
  pages={25261--25270},
  year={2025}
}

@inproceedings{qiu2024federated,
  title={Federated text-driven prompt generation for vision-language models},
  author={Qiu, Chen and Li, Xingyu and Mummadi, Chaithanya Kumar and Ganesh, Madan and Li, Zhenzhen and Peng, Lu and Lin, Wan-Yi},
  booktitle={International conference on learning representations},
  volume={2024},
  pages={12997--13012},
  year={2024}
}

@inproceedings{singha2025fedmvp,
  title={FedMVP: Federated Multimodal Visual Prompt Tuning for Vision-Language Models},
  author={Singha, Mainak and Roy, Subhankar and Mehrotra, Sarthak and Jha, Ankit and Abdar, Moloud and Banerjee, Biplab and Ricci, Elisa},
  booktitle={Proceedings of the IEEE/CVF International Conference on Computer Vision},
  pages={17869--17878},
  year={2025}
}

@inproceedings{wang2026fedmpt,
  title={FedMPT: Federated Multi-Label Prompt Tuning of Vision-Language Models},
  author={Wang, Xucong and Wang, Pengkun and Zhao, Zhe and Yu, Liheng and Wang, Shuang and Wang, Yang},
  booktitle={Proceedings of the IEEE/CVF Conference on Computer Vision and Pattern Recognition},
  pages={17226--17236},
  year={2026}
}

@inproceedings{singha2026biovlm,
  title={BioVLM: Routing Prompts, Not Parameters, for Cross-Modality Generalization in Biomedical VLMs},
  author={Singha, Mainak and Gupta, Tanisha and Jha, Ankit and Khan, Muhammad Haris and Ghosh, Sayantani and Banerjee, Biplab},
  booktitle={Findings of the Association for Computational Linguistics: ACL 2026},
  pages={40986--41005},
  year={2026}
}

@inproceedings{parthasarathy2026vimclip,
  title={VimCLIP: A Vision Mamba Based Multimodal Approach for Retrieval and Zero-Shot Classification Tasks},
  author={Parthasarathy, Ambarish and Channappayya, Sumohana},
  booktitle={2026 National Conference on Communications (NCC)},
  pages={460--465},
  year={2026},
  organization={IEEE}
}
\bibliographystyle{tmlr}
\newpage
\appendix

\section{Appendix}

\section{Notation}
\label{app:notation}
  
Table~\ref{tab:notation} summarizes the notation used throughout 
the paper.
\begin{table}[h!]
\centering
\caption{Summary of notation used in FedSSMCoOp.}
\label{tab:notation}
\footnotesize 
\begin{tabular}{ll}
\toprule
\textbf{Symbol} & \textbf{Description} \\
\midrule
\multicolumn{2}{l}{\textit{Federated setup}} \\
\midrule
$K$              & Number of clients (hospitals) \\
$k$              & Client index, $k \in \{1,\dots,K\}$ \\
$t$              & Communication round index \\
$T$              & Total number of communication rounds \\
$E$              & Number of local epochs per round \\
$\mathcal{D}_k$  & Local dataset of client $k$ \\
$n_k$            & Number of training samples at client $k$ \\
$\alpha$         & Dirichlet concentration parameter (data heterogeneity) \\
$C$              & Number of classes in the dataset \\
$k$              & Number of shots per class (few-shot setting) \\
\midrule
\multicolumn{2}{l}{\textit{Prompt learning}} \\
\midrule
$X_p$            & Learnable soft prompt (context tokens) \\
$X_p^{k}$        & Client $k$'s prompt after local training \\
$\bar{P}^{(t)}$  & Global prompt at round $t$ (server) \\
$\Delta X_p^{k}$ & Client $k$'s prompt update, $X_p^{k}-\bar{P}^{(t)}$ \\
$M$              & Number of context tokens ($M{=}4$) \\
$d$              & Embedding dimension of a prompt token ($d{=}512$) \\
$y$              & Ground-truth class label \\
\midrule
\multicolumn{2}{l}{\textit{Encoders and features}} \\
\midrule
$\mathcal{E}_v$  & Frozen Vision Mamba image encoder (Vim-S) \\
$\mathcal{E}_t$  & Frozen BiomedCLIP text encoder \\
$x$              & Input image \\
$V$              & Image tokens from $\mathcal{E}_v$ \\
$T$              & Text tokens from $\mathcal{E}_t$ \\
$V'$             & Text-aware image features (MIB output) \\
$T'$             & Image-aware text features (MIB output) \\
$\bar{V}'$       & Mean-pooled image features \\
$W_{txt}$        & Learnable text projection \\
$\tau$           & Temperature scaling parameter \\
\midrule
\multicolumn{2}{l}{\textit{Mamba Interaction Block (MIB)}} \\
\midrule
$L$              & Number of MIB layers \\
$r$              & Low-rank ratio in the MIB \\
$d_{v}$          & Image feature dimension in the MIB ($d_v{=}384$) \\
\midrule
\multicolumn{2}{l}{\textit{Objectives}} \\
\midrule
$\text{Pr}$      & Predicted class probability distribution \\
$\mathcal{L}_{CE}$ & Cross-entropy classification loss \\
\midrule
\multicolumn{2}{l}{\textit{LLM ablation only (not in the main method)}} \\
\midrule
$P_g$            & LLM-derived class prototype bank \\
$\mathcal{L}_{FedSCCM}$ & Semantic alignment loss, $\text{MSE}(T, P_g)$ \\
$\mathcal{L}_{FedKD}$   & Distillation loss, $\text{KL}(\text{Pr}_{teacher}\,\|\,\text{Pr}_{student})$ \\
$\lambda_{S}$    & Weight of the alignment loss \\
$\lambda_{K}$    & Weight of the distillation loss \\
$\alpha_{fuse}$  & Learnable prompt--prototype fusion coefficient \\
\bottomrule
\end{tabular}
\end{table}

\subsection{Few-shot Results Across Shot Settings}
We report the complete per-dataset few-shot accuracy for $k \in \{1, 2, 4, 8, 16\}$
shots in Tables~\ref{tab:main_results}--\ref{tab:main_results_16shot_transpose},
and the zero-shot results in Table~\ref{tab:zeroshot_results}. The results reveal a clear improvement in the performance of FedSSMCoOp as the number of labeled examples increases. At 1-shot, FedSSMCoOp achieves the best performance on 4 of the 11 datasets. This increases to 7 datasets at both 2- and 4-shot, 8 datasets at 8-shot, and 10 datasets at 16-shot. These results indicate that the proposed approach benefits consistently from additional labeled examples and performs particularly well when more supervision is available.

FedSSMCoOp achieves the best performance across all shot settings on CHMNIST, COVID, and LungColon. These datasets involve visual patterns that can benefit from capturing broader image-level context, which is consistent with the global feature modeling capability of the Vision Mamba encoder. For example, on LungColon, the accuracy improves from 68.6\% at 1-shot to 92.4\% at 8-shot.

However, the performance is less consistent on datasets that require fine-grained or localized visual discrimination. In particular, KneeXray does not achieve the best performance at any shot level, while BUSI, CTKidney, and DermaMNIST show larger performance gaps at lower shot levels. These results suggest that the global feature modeling of the SSM-based encoder may be less advantageous for tasks where subtle local differences or lesion-level details are important.

\begin{table*}[htp!]
\centering
\caption{Top-1 accuracy (\%) across all datasets and methods for zero-shot evaluation. \textbf{Best} and \textcolor{blue!70!black}{\underline{second-best}} results are highlighted for each dataset.}
\label{tab:zeroshot_results}
\tiny
\resizebox{\textwidth}{!}{%
\begin{tabular}{lcccc||c}
\toprule
\textbf{Dataset}
& \textbf{FedBiomedCLIP}
& \textbf{FedCoCoOp}
& \textbf{FedKgCoOp}
& \textbf{FedBiomedCoOp}
& \textbf{FedSSMCoOp (Ours)} \\
\midrule

\textbf{BTMRI}
& \textbf{57.08}
& 26.42{\scalebox{0.55}{$\pm$4.40}}
& 23.30
& \textcolor{blue!70!black}{\underline{31.00}}
& 23.74{\scalebox{0.55}{$\pm$2.67}} \\

\textbf{BUSI}
& \textbf{51.27}
& \textcolor{blue!70!black}{\underline{37.85{\scalebox{0.55}{$\pm$13.23}}}}
& 17.37
& 22.10
& 36.44{\scalebox{0.55}{$\pm$13.15}} \\

\textbf{CHMNIST}
& \textbf{29.06}
& 22.07{\scalebox{0.55}{$\pm$1.95}}
& \textcolor{blue!70!black}{\underline{23.14}}
& 22.85
& 17.87{\scalebox{0.55}{$\pm$1.75}} \\

\textbf{COVID}
& \textbf{45.84}
& 25.29{\scalebox{0.55}{$\pm$7.30}}
& \textcolor{blue!70!black}{\underline{31.29}}
& 29.40
& 30.23{\scalebox{0.55}{$\pm$8.98}} \\

\textbf{LungColon}
& \textbf{51.20}
& 22.97{\scalebox{0.55}{$\pm$3.69}}
& \textcolor{blue!70!black}{\underline{28.80}}
& 26.15
& 20.80{\scalebox{0.55}{$\pm$6.52}} \\

\textbf{OCTMNIST}
& \textbf{35.00}
& \textcolor{blue!70!black}{\underline{25.40{\scalebox{0.55}{$\pm$0.86}}}}
& 21.80
& 22.45
& 19.03{\scalebox{0.55}{$\pm$5.72}} \\

\textbf{DermaMNIST}
& \textcolor{blue!70!black}{\underline{30.87}}
& 11.20{\scalebox{0.55}{$\pm$6.87}}
& 12.87
& 12.10
& \textbf{43.71{\scalebox{0.55}{$\pm$13.78}}} \\

\textbf{RetinaMNIST}
& \textbf{27.68}
& 24.92{\scalebox{0.55}{$\pm$3.97}}
& \textcolor{blue!70!black}{\underline{26.10}}
& 25.50
& 24.92{\scalebox{0.55}{$\pm$3.97}} \\

\textbf{Kvasir}
& \textbf{55.50}
& 16.19{\scalebox{0.55}{$\pm$3.79}}
& \textcolor{blue!70!black}{\underline{18.25}}
& 17.40
& 18.03{\scalebox{0.55}{$\pm$1.84}} \\

\textbf{CTKidney}
& \textcolor{blue!70!black}{\underline{36.73}}
& 21.87{\scalebox{0.55}{$\pm$5.30}}
& \textbf{40.40}
& 34.80
& 21.33{\scalebox{0.55}{$\pm$5.98}} \\

\textbf{KneeXray}
& \textbf{27.48}
& 20.21{\scalebox{0.55}{$\pm$6.42}}
& \textcolor{blue!70!black}{\underline{25.60}}
& 23.10
& 19.51{\scalebox{0.55}{$\pm$5.16}} \\
\bottomrule
\end{tabular}
}
\end{table*}

\begin{table*}[htp!]
\centering
\caption{The quantitative few-shot classification accuracy ($k=1$) results on the respective baselines and the proposed framework on different datasets. For each of the datasets, \textbf{Best performing} and \textcolor{blue!70!black}{\underline{second best performing}} models are reported.}
\label{tab:main_results}
\tiny
\resizebox{\textwidth}{!}{%
\begin{tabular}{lccccccc||c}
\toprule
\textbf{Dataset} 
& \textbf{FedBiomedCLIP} 
& \textbf{FedCoCoOp} 
& \textbf{FedKgCoOp} 
& \textbf{PromptFL} 
& \textbf{PromptFL+FedProx} 
& \textbf{FedOTP} 
& \textbf{FedBiomedCoOp} 
& \textbf{FedSSMCoOp (Ours)} \\
\midrule

BTMRI 
& \textbf{57.08{\scalebox{0.55}{$\pm$0.00}}}
& 48.86{\scalebox{0.55}{$\pm$7.92}}
& 51.19{\scalebox{0.55}{$\pm$4.62}}
& 41.60{\scalebox{0.55}{$\pm$3.29}}
& 36.93{\scalebox{0.55}{$\pm$5.44}}
& 31.43{\scalebox{0.55}{$\pm$3.93}}
& 50.67{\scalebox{0.55}{$\pm$4.74}}
& \textcolor{blue!70!black}{\underline{53.64{\scalebox{0.55}{$\pm$1.78}}}} \\

BUSI \cite{al2020dataset}
& \textbf{51.27{\scalebox{0.55}{$\pm$0.00}}}
& 43.22{\scalebox{0.55}{$\pm$6.27}}
& 39.69{\scalebox{0.55}{$\pm$4.73}}
& 43.30{\scalebox{0.55}{$\pm$12.72}}
& 37.43{\scalebox{0.55}{$\pm$12.65}}
& 33.33{\scalebox{0.55}{$\pm$6.91}}
& \textcolor{blue!70!black}{\underline{46.61{\scalebox{0.55}{$\pm$7.79}}}}
& 40.11{\scalebox{0.55}{$\pm$6.75}} \\

CHMNIST \cite{kather2016multi}
& 29.06{\scalebox{0.55}{$\pm$0.00}}
& 49.53{\scalebox{0.55}{$\pm$1.90}}
& 45.48{\scalebox{0.55}{$\pm$2.53}}
& 30.57{\scalebox{0.55}{$\pm$6.17}}
& 33.60{\scalebox{0.55}{$\pm$3.78}}
& 34.06{\scalebox{0.55}{$\pm$9.48}}
& \textcolor{blue!70!black}{\underline{52.87{\scalebox{0.55}{$\pm$2.43}}}}
& \textbf{53.39{\scalebox{0.55}{$\pm$1.67}}} \\

COVID \cite{tahir2021covid}
& \textcolor{blue!70!black}{\underline{45.84{\scalebox{0.55}{$\pm$0.00}}}}
& 44.69{\scalebox{0.55}{$\pm$1.03}}
& 43.52{\scalebox{0.55}{$\pm$11.06}}
& 42.35{\scalebox{0.55}{$\pm$8.13}}
& 42.30{\scalebox{0.55}{$\pm$8.20}}
& 37.50{\scalebox{0.55}{$\pm$2.70}}
& 43.59{\scalebox{0.55}{$\pm$16.49}}
& \textbf{46.71{\scalebox{0.55}{$\pm$10.13}}} \\

LC25000 
& 51.20{\scalebox{0.55}{$\pm$0.00}}
& 58.72{\scalebox{0.55}{$\pm$2.63}}
& 55.99{\scalebox{0.55}{$\pm$7.84}}
& 42.90{\scalebox{0.55}{$\pm$5.52}}
& 55.50{\scalebox{0.55}{$\pm$5.33}}
& 26.25{\scalebox{0.55}{$\pm$8.40}}
& \textcolor{blue!70!black}{\underline{61.88{\scalebox{0.55}{$\pm$2.99}}}}
& \textbf{68.61{\scalebox{0.55}{$\pm$4.67}}} \\

OCTMNIST 
& 35.00{\scalebox{0.55}{$\pm$0.00}}
& \textbf{43.30{\scalebox{0.55}{$\pm$2.18}}}
& 32.43{\scalebox{0.55}{$\pm$2.00}}
& \textcolor{blue!70!black}{\underline{41.80{\scalebox{0.55}{$\pm$9.72}}}}
& 40.65{\scalebox{0.55}{$\pm$9.55}}
& 26.25{\scalebox{0.55}{$\pm$8.40}}
& 32.57{\scalebox{0.55}{$\pm$9.67}}
& 34.93{\scalebox{0.55}{$\pm$5.76}} \\

DermaMNIST 
& \textbf{30.87{\scalebox{0.55}{$\pm$0.00}}}
& 19.63{\scalebox{0.55}{$\pm$3.27}}
& 16.99{\scalebox{0.55}{$\pm$1.25}}
& 21.20{\scalebox{0.55}{$\pm$15.79}}
& 18.20{\scalebox{0.55}{$\pm$7.34}}
& 14.66{\scalebox{0.55}{$\pm$3.09}}
& 15.73{\scalebox{0.55}{$\pm$1.71}}
& \textcolor{blue!70!black}{\underline{22.72{\scalebox{0.55}{$\pm$3.00}}}} \\

RetinaMNIST 
& 27.68{\scalebox{0.55}{$\pm$0.00}}
& 33.81{\scalebox{0.55}{$\pm$5.17}}
& 39.36{\scalebox{0.55}{$\pm$6.29}}
& 28.43{\scalebox{0.55}{$\pm$1.81}}
& 30.60{\scalebox{0.55}{$\pm$4.05}}
& 28.75{\scalebox{0.55}{$\pm$4.53}}
& \textcolor{blue!70!black}{\underline{40.25{\scalebox{0.55}{$\pm$6.32}}}}
& \textbf{42.98{\scalebox{0.55}{$\pm$2.25}}} \\

Kvasir 
& \textbf{55.50{\scalebox{0.55}{$\pm$0.00}}}
& 42.97{\scalebox{0.55}{$\pm$1.12}}
& 39.72{\scalebox{0.55}{$\pm$2.19}}
& 28.90{\scalebox{0.55}{$\pm$6.37}}
& 26.33{\scalebox{0.55}{$\pm$2.99}}
& 19.97{\scalebox{0.55}{$\pm$2.53}}
& 42.94{\scalebox{0.55}{$\pm$3.95}}
& \textcolor{blue!70!black}{\underline{46.80{\scalebox{0.55}{$\pm$2.79}}}} \\

CTKidney 
& 36.73{\scalebox{0.55}{$\pm$0.00}}
& 31.96{\scalebox{0.55}{$\pm$4.00}}
& \textcolor{blue!70!black}{\underline{37.46{\scalebox{0.55}{$\pm$7.72}}}}
& \textbf{42.98{\scalebox{0.55}{$\pm$2.25}}}
& 32.19{\scalebox{0.55}{$\pm$3.82}}
& 27.56{\scalebox{0.55}{$\pm$13.87}}
& 32.75{\scalebox{0.55}{$\pm$3.17}}
& 29.03{\scalebox{0.55}{$\pm$6.35}} \\

KneeXray \cite{chen2018knee}
& 27.48{\scalebox{0.55}{$\pm$0.00}}
& 31.62{\scalebox{0.55}{$\pm$5.73}}
& \textcolor{blue!70!black}{\underline{34.42{\scalebox{0.55}{$\pm$2.92}}}}
& 24.83{\scalebox{0.55}{$\pm$7.42}}
& 30.25{\scalebox{0.55}{$\pm$9.78}}
& 19.10{\scalebox{0.55}{$\pm$6.73}}
& \textbf{38.30{\scalebox{0.55}{$\pm$1.18}}}
& 26.67{\scalebox{0.55}{$\pm$2.30}} \\

\bottomrule
\end{tabular}
}
\end{table*}
\begin{table*}[htp!]
\centering
\caption{Few-shot classification accuracy (\%) under the 2-shot setting across biomedical datasets.For each of the datasets, \textbf{Best performing} and \textcolor{blue!70!black}{\underline{second best performing}} models are reported.}
\label{tab:main_results_transpose}
\tiny
\resizebox{\textwidth}{!}{%
\begin{tabular}{lcccccc}
\toprule
\textbf{Dataset} &

\textbf{FedCoCoOp} &
\textbf{FedKgCoOp} &
\textbf{PromptFL} &
\textbf{PromptFL+FedProx} &
\textbf{FedOTP} &
\textbf{FedSSMCoOp (Ours)} \\
\midrule

\textbf{BTMRI}
& 49.80{\scalebox{0.55}{$\pm$3.56}}
& \textcolor{blue!70!black}{\underline{53.37{\scalebox{0.55}{$\pm$6.67}}}}
& 37.20{\scalebox{0.55}{$\pm$6.41}}
& 34.20{\scalebox{0.55}{$\pm$5.38}}
& 39.42{\scalebox{0.55}{$\pm$11.99}}
& \textbf{57.40{\scalebox{0.55}{$\pm$4.59}}} \\

\textbf{BUSI}

& \textcolor{blue!70!black}{\underline{48.31{\scalebox{0.55}{$\pm$3.00}}}}
& 42.66{\scalebox{0.55}{$\pm$7.50}}
& 32.43{\scalebox{0.55}{$\pm$7.02}}
& 26.80
& \textbf{57.73{\scalebox{0.55}{$\pm$0.24}}}
& 45.48{\scalebox{0.55}{$\pm$3.21}} \\

\textbf{CHMNIST}
& \textcolor{blue!70!black}{\underline{53.08{\scalebox{0.55}{$\pm$1.51}}}}
& 44.99{\scalebox{0.55}{$\pm$3.10}}
& 26.27{\scalebox{0.55}{$\pm$12.55}}
& 32.93{\scalebox{0.55}{$\pm$5.60}}
& 23.40{\scalebox{0.55}{$\pm$5.15}}
& \textbf{60.61{\scalebox{0.55}{$\pm$2.68}}} \\

\textbf{COVID}

& \textcolor{blue!70!black}{\underline{41.78{\scalebox{0.55}{$\pm$11.49}}}}
& 28.72{\scalebox{0.55}{$\pm$6.34}}
& 20.90{\scalebox{0.55}{$\pm$7.79}}
& 37.63{\scalebox{0.55}{$\pm$18.39}}
& 26.32{\scalebox{0.55}{$\pm$14.12}}
& \textbf{54.17{\scalebox{0.55}{$\pm$7.22}}} \\

\textbf{LungColon}

& \textcolor{blue!70!black}{\underline{53.17{\scalebox{0.55}{$\pm$12.04}}}}
& 46.44{\scalebox{0.55}{$\pm$6.89}}
& 46.44{\scalebox{0.55}{$\pm$6.89}}
& 45.90{\scalebox{0.55}{$\pm$4.92}}
& 40.81{\scalebox{0.55}{$\pm$16.30}}
& \textbf{79.09{\scalebox{0.55}{$\pm$6.20}}} \\

\textbf{OCTMNIST}
& \textcolor{blue!70!black}{\underline{37.33{\scalebox{0.55}{$\pm$5.00}}}}
& 33.10{\scalebox{0.55}{$\pm$2.45}}
& 31.40{\scalebox{0.55}{$\pm$6.10}}
& 31.57{\scalebox{0.55}{$\pm$9.67}}
& \textbf{40.00{\scalebox{0.55}{$\pm$4.81}}}
& 32.50{\scalebox{0.55}{$\pm$5.80}} \\

\textbf{DermaMNIST}
& 19.82{\scalebox{0.55}{$\pm$2.94}}
& \textcolor{blue!70!black}{\underline{24.32{\scalebox{0.55}{$\pm$18.52}}}}
& 23.43{\scalebox{0.55}{$\pm$12.39}}
& 15.40{\scalebox{0.55}{$\pm$14.54}}
& 9.14{\scalebox{0.55}{$\pm$4.03}}
& \textbf{29.96{\scalebox{0.55}{$\pm$6.15}}} \\

\textbf{RetinaMNIST}
& 43.77{\scalebox{0.55}{$\pm$2.86}}
& \textbf{46.01{\scalebox{0.55}{$\pm$4.78}}}
& 27.27{\scalebox{0.55}{$\pm$6.69}}
& 31.43{\scalebox{0.55}{$\pm$2.48}}
& 34.89{\scalebox{0.55}{$\pm$9.83}}
& \textcolor{blue!70!black}{\underline{45.19{\scalebox{0.55}{$\pm$6.56}}}} \\

\textbf{Kvasir}

& \textcolor{blue!70!black}{\underline{44.33{\scalebox{0.55}{$\pm$5.72}}}}
& 30.08{\scalebox{0.55}{$\pm$1.89}}
& 28.90{\scalebox{0.55}{$\pm$1.04}}
& 27.47{\scalebox{0.55}{$\pm$1.23}}
& 24.73{\scalebox{0.55}{$\pm$4.07}}
& \textbf{57.39{\scalebox{0.55}{$\pm$3.50}}} \\

\textbf{CTKidney}
& 39.49{\scalebox{0.55}{$\pm$7.31}}
& \textcolor{blue!70!black}{\underline{43.01{\scalebox{0.55}{$\pm$3.20}}}}
& 40.00{\scalebox{0.55}{$\pm$5.29}}
& 40.00{\scalebox{0.55}{$\pm$4.32}}
& 23.46{\scalebox{0.55}{$\pm$6.31}}
& \textbf{44.11{\scalebox{0.55}{$\pm$4.44}}} \\

\textbf{KneeXray}

& \textbf{37.78{\scalebox{0.55}{$\pm$0.51}}}
& \textcolor{blue!70!black}{\underline{29.43{\scalebox{0.55}{$\pm$5.04}}}}
& 24.70{\scalebox{0.55}{$\pm$7.28}}
& 28.90{\scalebox{0.55}{$\pm$8.20}}
& 24.20{\scalebox{0.55}{$\pm$3.48}}
& 22.16{\scalebox{0.55}{$\pm$2.27}} \\

\bottomrule
\end{tabular}
}
\end{table*}
\begin{table*}[htp!]
\centering
\caption{Few-shot classification accuracy (\%) under the 4-shot setting across biomedical datasets. For each of the datasets, \textbf{Best performing} and \textcolor{blue!70!black}{\underline{second best performing}} models are reported.}
\label{tab:main_results_4shot_transpose}
\tiny
\resizebox{\textwidth}{!}{%
\begin{tabular}{lcccccc}
\toprule
\textbf{Dataset}

& \textbf{FedCoCoOp}
& \textbf{FedKgCoOp}
& \textbf{PromptFL}
& \textbf{PromptFL+FedProx}
& \textbf{FedOTP}
& \textbf{FedSSMCoOp (Ours)} \\
\midrule

\textbf{BTMRI}

& \textcolor{blue!70!black}{\underline{51.20{\scalebox{0.55}{$\pm$4.10}}}}
& 48.71{\scalebox{0.55}{$\pm$0.32}}
& 46.80{\scalebox{0.55}{$\pm$4.00}}
& 32.23{\scalebox{0.55}{$\pm$5.86}}
& 46.67{\scalebox{0.55}{$\pm$12.04}}
& \textbf{60.12{\scalebox{0.55}{$\pm$2.07}}} \\

\textbf{BUSI}

& 47.46{\scalebox{0.55}{$\pm$15.16}}
& 46.33{\scalebox{0.55}{$\pm$2.34}}
& \textbf{56.43{\scalebox{0.55}{$\pm$0.71}}}
& \textcolor{blue!70!black}{\underline{55.55{\scalebox{0.55}{$\pm$2.75}}}}
& 39.30{\scalebox{0.55}{$\pm$12.10}}
& 55.08{\scalebox{0.55}{$\pm$5.96}} \\

\textbf{CHMNIST}

& 46.65{\scalebox{0.55}{$\pm$9.26}}
& 40.43{\scalebox{0.55}{$\pm$6.76}}
& 36.33{\scalebox{0.55}{$\pm$3.68}}
& 33.70{\scalebox{0.55}{$\pm$0.86}}
& 37.33{\scalebox{0.55}{$\pm$3.07}}
& \textbf{71.23{\scalebox{0.55}{$\pm$2.84}}} \\

\textbf{COVID}

& 37.14{\scalebox{0.55}{$\pm$5.31}}
& 40.56{\scalebox{0.55}{$\pm$13.44}}
& \textcolor{blue!70!black}{\underline{48.40{\scalebox{0.55}{$\pm$0.71}}}}
& 48.10{\scalebox{0.55}{$\pm$1.66}}
& 27.10{\scalebox{0.55}{$\pm$8.49}}
& \textbf{53.81{\scalebox{0.55}{$\pm$8.96}}} \\

\textbf{LungColon}

& \textcolor{blue!70!black}{\underline{55.00{\scalebox{0.55}{$\pm$2.88}}}}
& 44.65{\scalebox{0.55}{$\pm$6.88}}
& 48.27{\scalebox{0.55}{$\pm$0.88}}
& 46.50{\scalebox{0.55}{$\pm$4.15}}
& 52.80{\scalebox{0.55}{$\pm$2.49}}
& \textbf{88.24{\scalebox{0.55}{$\pm$1.67}}} \\

\textbf{OCTMNIST}

& \textcolor{blue!70!black}{\underline{47.27{\scalebox{0.55}{$\pm$7.45}}}}
& 41.35{\scalebox{0.55}{$\pm$3.10}}
& 38.37{\scalebox{0.55}{$\pm$5.68}}
& 31.23{\scalebox{0.55}{$\pm$14.20}}
& 36.53{\scalebox{0.55}{$\pm$14.48}}
& \textbf{48.03{\scalebox{0.55}{$\pm$1.90}}} \\

\textbf{DermaMNIST}

& 23.66{\scalebox{0.55}{$\pm$6.25}}
& 15.69{\scalebox{0.55}{$\pm$4.32}}
& \textbf{42.90{\scalebox{0.55}{$\pm$11.30}}}
& 8.40{\scalebox{0.55}{$\pm$4.19}}
& 18.51{\scalebox{0.55}{$\pm$13.81}}
& \textcolor{blue!70!black}{\underline{39.54{\scalebox{0.55}{$\pm$10.68}}}} \\

\textbf{RetinaMNIST}

& \textcolor{blue!70!black}{\underline{41.90{\scalebox{0.55}{$\pm$5.70}}}}
& 39.36{\scalebox{0.55}{$\pm$6.29}}
& 32.13{\scalebox{0.55}{$\pm$5.39}}
& 32.37{\scalebox{0.55}{$\pm$4.92}}
& 38.30{\scalebox{0.55}{$\pm$8.56}}
& \textbf{53.29{\scalebox{0.55}{$\pm$3.13}}} \\

\textbf{Kvasir}

& \textcolor{blue!70!black}{\underline{45.28{\scalebox{0.55}{$\pm$1.41}}}}
& 35.08{\scalebox{0.55}{$\pm$0.98}}
& 30.33{\scalebox{0.55}{$\pm$5.64}}
& 28.95{\scalebox{0.55}{$\pm$4.30}}
& 41.14{\scalebox{0.55}{$\pm$10.82}}
& \textbf{67.33{\scalebox{0.55}{$\pm$1.06}}} \\

\textbf{CTKidney}

& 40.36{\scalebox{0.55}{$\pm$7.56}}
& \textcolor{blue!70!black}{\underline{40.62{\scalebox{0.55}{$\pm$1.80}}}}
& \textbf{48.58{\scalebox{0.55}{$\pm$2.88}}}
& 42.10{\scalebox{0.55}{$\pm$3.25}}
& 34.64{\scalebox{0.55}{$\pm$7.88}}
& 44.13{\scalebox{0.55}{$\pm$3.38}} \\

\textbf{KneeXray}

& 29.83{\scalebox{0.55}{$\pm$3.88}}
& \textbf{35.30{\scalebox{0.55}{$\pm$3.54}}}
& \textcolor{blue!70!black}{\underline{33.33{\scalebox{0.55}{$\pm$8.02}}}}
& 27.15{\scalebox{0.55}{$\pm$5.40}}
& 24.20{\scalebox{0.55}{$\pm$3.48}}
& 27.57{\scalebox{0.55}{$\pm$1.13}} \\

\bottomrule
\end{tabular}
}
\end{table*}
\begin{table*}[htp!]
\centering
\caption{Few-shot classification accuracy (\%) under the 8-shot setting across biomedical datasets. For each of the datasets, \textbf{Best performing} and \textcolor{blue!70!black}{\underline{second best performing}} models are reported.}
\label{tab:main_results_8shot_transpose}
\tiny
\resizebox{\textwidth}{!}{%
\begin{tabular}{lcccccc}
\toprule
\textbf{Dataset}

& \textbf{FedCoCoOp}
& \textbf{FedKgCoOp}
& \textbf{PromptFL}
& \textbf{PromptFL+FedProx}
& \textbf{FedOTP}
& \textbf{FedSSMCoOp (Ours)} \\
\midrule

\textbf{BTMRI}

& \textcolor{blue!70!black}{\underline{55.12{\scalebox{0.55}{$\pm$4.20}}}}
& 52.28{\scalebox{0.55}{$\pm$4.14}}
& 48.00{\scalebox{0.55}{$\pm$3.77}}
& 30.03{\scalebox{0.55}{$\pm$5.95}}
& 41.83{\scalebox{0.55}{$\pm$4.10}}
& \textbf{73.62{\scalebox{0.55}{$\pm$1.64}}} \\

\textbf{BUSI}

& \textcolor{blue!70!black}{\underline{49.43{\scalebox{0.55}{$\pm$11.53}}}}
& 40.68{\scalebox{0.55}{$\pm$14.56}}
& 46.50{\scalebox{0.55}{$\pm$8.20}}
& 43.10{\scalebox{0.55}{$\pm$6.45}}
& 36.47{\scalebox{0.55}{$\pm$8.73}}
& \textbf{61.30{\scalebox{0.55}{$\pm$8.86}}} \\

\textbf{CHMNIST}

& \textcolor{blue!70!black}{\underline{57.45{\scalebox{0.55}{$\pm$5.08}}}}
& 41.29{\scalebox{0.55}{$\pm$1.05}}
& 42.10{\scalebox{0.55}{$\pm$4.90}}
& 39.25{\scalebox{0.55}{$\pm$3.10}}
& 42.47{\scalebox{0.55}{$\pm$7.77}}
& \textbf{78.01{\scalebox{0.55}{$\pm$0.60}}} \\

\textbf{COVID}

& \textcolor{blue!70!black}{\underline{52.69{\scalebox{0.55}{$\pm$6.43}}}}
& 42.59{\scalebox{0.55}{$\pm$8.91}}
& 46.80{\scalebox{0.55}{$\pm$5.15}}
& 44.20{\scalebox{0.55}{$\pm$2.80}}
& 31.97{\scalebox{0.55}{$\pm$12.54}}
& \textbf{64.03{\scalebox{0.55}{$\pm$4.18}}} \\

\textbf{LungColon}
& \textcolor{blue!70!black}{\underline{65.01{\scalebox{0.55}{$\pm$4.54}}}}
& 50.32{\scalebox{0.55}{$\pm$4.11}}
& 51.15{\scalebox{0.55}{$\pm$3.20}}
& 48.40{\scalebox{0.55}{$\pm$4.10}}
& 44.77{\scalebox{0.55}{$\pm$4.98}}
& \textbf{92.40{\scalebox{0.55}{$\pm$1.10}}} \\

\textbf{OCTMNIST}

& \textcolor{blue!70!black}{\underline{47.47{\scalebox{0.55}{$\pm$3.40}}}}
& 44.25{\scalebox{0.55}{$\pm$2.90}}
& 40.10{\scalebox{0.55}{$\pm$5.10}}
& 34.80{\scalebox{0.55}{$\pm$9.15}}
& 32.80{\scalebox{0.55}{$\pm$12.95}}
& \textbf{61.20{\scalebox{0.55}{$\pm$1.70}}} \\

\textbf{DermaMNIST}

& 28.30{\scalebox{0.55}{$\pm$4.40}}
& 12.80{\scalebox{0.55}{$\pm$1.14}}
& \textbf{50.70{\scalebox{0.55}{$\pm$2.78}}}
& 24.10{\scalebox{0.55}{$\pm$16.23}}
& 23.60{\scalebox{0.55}{$\pm$20.50}}
& \textcolor{blue!70!black}{\underline{42.93{\scalebox{0.55}{$\pm$6.31}}}} \\

\textbf{RetinaMNIST}

& \textcolor{blue!70!black}{\underline{48.26{\scalebox{0.55}{$\pm$7.55}}}}
& 46.64{\scalebox{0.55}{$\pm$1.65}}
& 35.37{\scalebox{0.55}{$\pm$0.95}}
& 43.30{\scalebox{0.55}{$\pm$5.40}}
& 43.33{\scalebox{0.55}{$\pm$6.03}}
& \textbf{62.15{\scalebox{0.55}{$\pm$3.04}}} \\

\textbf{Kvasir}

& \textbf{53.50{\scalebox{0.55}{$\pm$4.19}}}
& 34.36{\scalebox{0.55}{$\pm$2.96}}
& 32.23{\scalebox{0.55}{$\pm$6.29}}
& 31.50{\scalebox{0.55}{$\pm$4.85}}
& 40.97{\scalebox{0.55}{$\pm$9.19}}
& \textcolor{blue!70!black}{\underline{43.03{\scalebox{0.55}{$\pm$8.40}}}} \\

\textbf{CTKidney}
& 39.72{\scalebox{0.55}{$\pm$3.99}}
& 38.58{\scalebox{0.55}{$\pm$2.27}}
& \textcolor{blue!70!black}{\underline{38.35{\scalebox{0.55}{$\pm$10.35}}}}
& 35.10{\scalebox{0.55}{$\pm$6.20}}
& 41.28{\scalebox{0.55}{$\pm$5.51}}
& \textbf{53.90{\scalebox{0.55}{$\pm$3.80}}} \\

\textbf{KneeXray}
& 26.87{\scalebox{0.55}{$\pm$13.49}}
& \textbf{32.45{\scalebox{0.55}{$\pm$5.46}}}
& 26.77{\scalebox{0.55}{$\pm$9.08}}
& 25.20{\scalebox{0.55}{$\pm$6.90}}
& 28.70{\scalebox{0.55}{$\pm$8.03}}
& \textcolor{blue!70!black}{\underline{32.20{\scalebox{0.55}{$\pm$3.70}}}} \\
\bottomrule
\end{tabular}
}
\end{table*}
\begin{table*}[htp!]
\centering
\caption{Few-shot classification accuracy (\%) under the 16-shot setting across biomedical datasets. For each of the datasets, \textbf{Best performing} and \textcolor{blue!70!black}{\underline{second best performing}} models are reported.}
\label{tab:main_results_16shot_transpose}
\tiny
\resizebox{\textwidth}{!}{%
\begin{tabular}{lcccccc}
\toprule
\textbf{Dataset}

& \textbf{FedCoCoOp}
& \textbf{FedKgCoOp}
& \textbf{PromptFL}
& \textbf{PromptFL+FedProx}
& \textbf{FedOTP}
& \textbf{FedSSMCoOp (Ours)} \\
\midrule

\textbf{BTMRI}

& \textcolor{blue!70!black}{\underline{60.86{\scalebox{0.55}{$\pm$5.00}}}}
& 53.21{\scalebox{0.55}{$\pm$3.14}}
& 52.60{\scalebox{0.55}{$\pm$5.11}}
& 37.00{\scalebox{0.55}{$\pm$8.65}}
& 48.23{\scalebox{0.55}{$\pm$6.49}}
& \textbf{76.06{\scalebox{0.55}{$\pm$1.31}}} \\

\textbf{BUSI}

\textbf{BUSI}
& \textcolor{blue!70!black}{\underline{60.87{\scalebox{0.55}{$\pm$6.20}}}}
& 51.70{\scalebox{0.55}{$\pm$5.34}}
& 50.15{\scalebox{0.55}{$\pm$7.40}}
& 47.50{\scalebox{0.55}{$\pm$5.10}}
& 40.30{\scalebox{0.55}{$\pm$10.28}}
& \textbf{69.21{\scalebox{0.55}{$\pm$5.04}}} \\

\textbf{CHMNIST}
& \textcolor{blue!70!black}{\underline{62.17{\scalebox{0.55}{$\pm$3.35}}}}
& 48.23{\scalebox{0.55}{$\pm$0.79}}
& 46.40{\scalebox{0.55}{$\pm$4.30}}
& 43.10{\scalebox{0.55}{$\pm$2.85}}
& 49.93{\scalebox{0.55}{$\pm$8.58}}
& \textbf{81.51{\scalebox{0.55}{$\pm$1.71}}} \\

\textbf{COVID}
& \textcolor{blue!70!black}{\underline{44.54{\scalebox{0.55}{$\pm$2.50}}}}
& 29.59{\scalebox{0.55}{$\pm$9.41}}
& 41.20{\scalebox{0.55}{$\pm$5.90}}
& 39.80{\scalebox{0.55}{$\pm$3.40}}
& 23.20{\scalebox{0.55}{$\pm$11.15}}
& \textbf{72.52{\scalebox{0.55}{$\pm$1.62}}} \\

\textbf{LungColon}
& \textcolor{blue!70!black}{\underline{73.71{\scalebox{0.55}{$\pm$4.01}}}}
& 52.30{\scalebox{0.55}{$\pm$3.90}}
& 54.80{\scalebox{0.55}{$\pm$2.15}}
& 51.20{\scalebox{0.55}{$\pm$3.50}}
& 55.67{\scalebox{0.55}{$\pm$5.09}}
& \textbf{92.97{\scalebox{0.55}{$\pm$1.32}}} \\

\textbf{OCTMNIST}
& \textcolor{blue!70!black}{\underline{57.13{\scalebox{0.55}{$\pm$8.04}}}}
& 47.10{\scalebox{0.55}{$\pm$3.15}}
& 43.40{\scalebox{0.55}{$\pm$5.25}}
& 38.50{\scalebox{0.55}{$\pm$8.60}}
& 39.87{\scalebox{0.55}{$\pm$16.34}}
& \textbf{57.60{\scalebox{0.55}{$\pm$3.95}}} \\

\textbf{DermaMNIST}
& 34.20{\scalebox{0.55}{$\pm$5.10}}
& 15.91{\scalebox{0.55}{$\pm$1.38}}
& 20.10{\scalebox{0.55}{$\pm$18.65}}
& \textcolor{blue!70!black}{\underline{37.13{\scalebox{0.55}{$\pm$25.87}}}}
& 19.97{\scalebox{0.55}{$\pm$1.13}}
& \textbf{45.02{\scalebox{0.55}{$\pm$6.15}}} \\

\textbf{RetinaMNIST}
& \textcolor{blue!70!black}{\underline{55.47{\scalebox{0.55}{$\pm$3.70}}}}
& 51.02{\scalebox{0.55}{$\pm$6.36}}
& 38.80{\scalebox{0.55}{$\pm$0.00}}
& 45.20{\scalebox{0.55}{$\pm$5.10}}
& 44.50{\scalebox{0.55}{$\pm$7.41}}
& \textbf{66.40{\scalebox{0.55}{$\pm$0.90}}} \\

\textbf{Kvasir}
& \textcolor{blue!70!black}{\underline{65.08{\scalebox{0.55}{$\pm$1.71}}}}
& 37.03{\scalebox{0.55}{$\pm$5.04}}
& 35.10{\scalebox{0.55}{$\pm$6.30}}
& 33.80{\scalebox{0.55}{$\pm$4.25}}
& 47.60{\scalebox{0.55}{$\pm$5.90}}
& \textbf{79.90{\scalebox{0.55}{$\pm$1.15}}} \\

\textbf{CTKidney}
& \textcolor{blue!70!black}{\underline{46.34{\scalebox{0.55}{$\pm$6.10}}}}
& 38.17{\scalebox{0.55}{$\pm$2.01}}
& 41.50{\scalebox{0.55}{$\pm$8.90}}
& 38.20{\scalebox{0.55}{$\pm$4.60}}
& 43.15{\scalebox{0.55}{$\pm$5.20}}
& \textbf{58.12{\scalebox{0.55}{$\pm$2.95}}} \\

\textbf{KneeXray}
& 29.10{\scalebox{0.55}{$\pm$8.45}}
& 26.49{\scalebox{0.55}{$\pm$4.85}}
& \textbf{33.03{\scalebox{0.55}{$\pm$5.35}}}
& 28.40{\scalebox{0.55}{$\pm$6.10}}
& 22.93{\scalebox{0.55}{$\pm$5.76}}
& \textcolor{blue!70!black}{\underline{31.45{\scalebox{0.55}{$\pm$3.10}}}} \\

\bottomrule
\end{tabular}
}
\end{table*}

\subsection{Dataset Characteristics}
\label{sec:dataset_char}
The results in Tables \ref{tab:main_results_transpose} \ref{tab:main_results_16shot_transpose} show that backbone performance varies with the characteristics of a dataset. Datasets characterized by global morphology or whole image context, such as CHMNIST, COVID, LC25000, and RetinaMNIST, generally favour Vision Mamba, with FedSSMCoOp performing best across most shot settings. In contrast, datasets requiring fine-grained or localized discrimination, such as BUSI and KneeXray, favor attention-based backbones. These observations are summarized in Table \ref{tab:dataset_analysis}
\begin{table}[t]
\centering
\caption{Observed relationship between dataset characteristics and best-performing backbone family, based on the few-shot results across shot settings.}
\label{tab:dataset_analysis}
\footnotesize
\setlength{\tabcolsep}{4pt}
\renewcommand{\arraystretch}{0.95}
\begin{tabular}{@{}lll@{}}
\toprule
\textbf{Dataset} & \textbf{Discriminative signal} & \textbf{Favoured backbone} \\
\midrule
CHMNIST     & Global morphology  & Vision Mamba \\
COVID       & Global context     & Vision Mamba \\
LC25000     & Tissue context     & Vision Mamba \\
RetinaMNIST & Global structures  & Vision Mamba \\
CTKidney    & Fine details       & Shot-dependent \\
BUSI        & Local lesions      & Attention \\
KneeXray    & Fine-grained grading & Attention \\
\bottomrule
\end{tabular}
\end{table}

\end{document}